\documentclass[12pt,letterpaper]{article}
\usepackage[a4paper, total={7in, 10in}]{geometry}

\usepackage{graphicx}

\usepackage{helvet}
\usepackage{authblk}
\usepackage{hyperref}
\usepackage{amsmath} 
\usepackage{amssymb} 
\usepackage{orcidlink} 
\usepackage{times}
\usepackage{latexsym}
\usepackage{colortbl}
\usepackage[x11names]{xcolor} 
\usepackage{svg}

\usepackage{booktabs}
\usepackage{makecell}
\usepackage{url}

\usepackage{multicol}
\usepackage{multirow}
\usepackage{marvosym}
\usepackage[super,comma,sort&compress]  
   {natbib}

\makeatletter
\@ifundefined{urlprefix}{}{}
\makeatother

\makeatletter
\renewcommand{\maketitle}{\bgroup\setlength{\parindent}{0pt}
\begin{flushleft}
  \textbf{\@title}
  
  \@author
\end{flushleft}\egroup}
\makeatother

\title{IoT-LLM: a framework for enhancing Large Language Model reasoning from real-world sensor data}
\date{}

\author[1]{Tuo An}
\author[2]{Yunjiao Zhou}
\author[1]{Han Zou}
\author[1,*,\Letter]{Jianfei Yang}

\affil[1]{MARS Lab, School of Mechanical and Aerospace Engineering, Nanyang Technological University, Singapore 639798}
\affil[2]{School of Electrical and Electronics Engineering, Nanyang Technological University, Singapore 639798}
\affil[*]{Correspondence: jianfei.yang@ntu.edu.sg}
\affil[\Letter]{Lead contact: jianfei.yang@ntu.edu.sg}

\begin{document}

\maketitle

\section*{SUMMARY}

Large Language Models (LLMs) excel in textual tasks but often struggle with physical-world reasoning tasks. Inspired by human cognition—where perception is fundamental to reasoning—we explore augmenting LLMs with enhanced perception abilities using Internet of Things (IoT) data and pertinent knowledge. In this work, we systematically study LLMs' capability to address IoT-sensory tasks by augmenting their perception and knowledge base, and then propose a unified framework, IoT-LLM, to enhance such capability. In IoT-LLM, we customize three steps: preprocessing IoT data into suitable formats, expanding LLMs knowledge via IoT-oriented retrieval-augmented generation and activating LLMs commonsense knowledge through chain-of-thought prompting. We design a benchmark comprising five real-world tasks with varying data types and reasoning complexities to evaluate the performance of IoT-LLM. Experimental results reveal that IoT-LLM significantly improves the performance of IoT-sensory task reasoning of LLMs, with models like GPT-4o-mini showing a 49.4\% average improvement over previous methods.

\section*{KEYWORDS}

Large language models, Physical AI, Agentic AI, Internet of Things, LLM Reasoning.


\section*{INTRODUCTION}

Recent advances in large generative models have showcased their exceptional performance and versatility in handling complex tasks across textual and visual domains, as evidenced by the GPT series~\citep{radford2018improving,radford2019language,brown2020language,achiam2023gpt} and visual generation models~\citep{dosovitskiy2021image,Ho2020DenoisingDP,peebles2023scalable,blattmann2023stable}. Given that current large language models (LLMs) possess extensive commonsense and world knowledge, they are promising candidates to serve as the core foundation for embodied AI tasks. In this context, enabling LLMs to accurately perceive and understand the physical world is fundamental to the future development of embodied AI and robotics. For example, by leveraging LLMs as base models and further training them with robotics data, it is possible to develop a unified Vision-Language-Action (VLA) model capable of handling a wide range of robotics tasks, such as daily household activities\citep{kim2024openvla,mu2023embodiedgpt,wang2024large}. However, these models could occasionally generate outputs that are physically implausible, often referred to as \textquotedblleft hallucinations\textquotedblright ~\citep{alkaissi2023artificial,huang2023survey}. Even advanced video generation models, e.g., Sora~\citep{videoworldsimulators2024}, are susceptible to producing animations that contravene fundamental physical laws, such as a video clip containing a tipping water glass that appears to defy gravity. These observations suggest that generative models may not really comprehend and apply physical laws of the physical world as accurately as humans when acting as world simulators. This has renewed interest in research on the \textit{World Model} that focuses on understanding and modeling the physical world in a brain-like manner~\citep{dawid2023introduction,garrido2024learning,liu2024world}.

\begin{figure*}[t!]
\begin{center}
\includegraphics[width=1.0\textwidth]{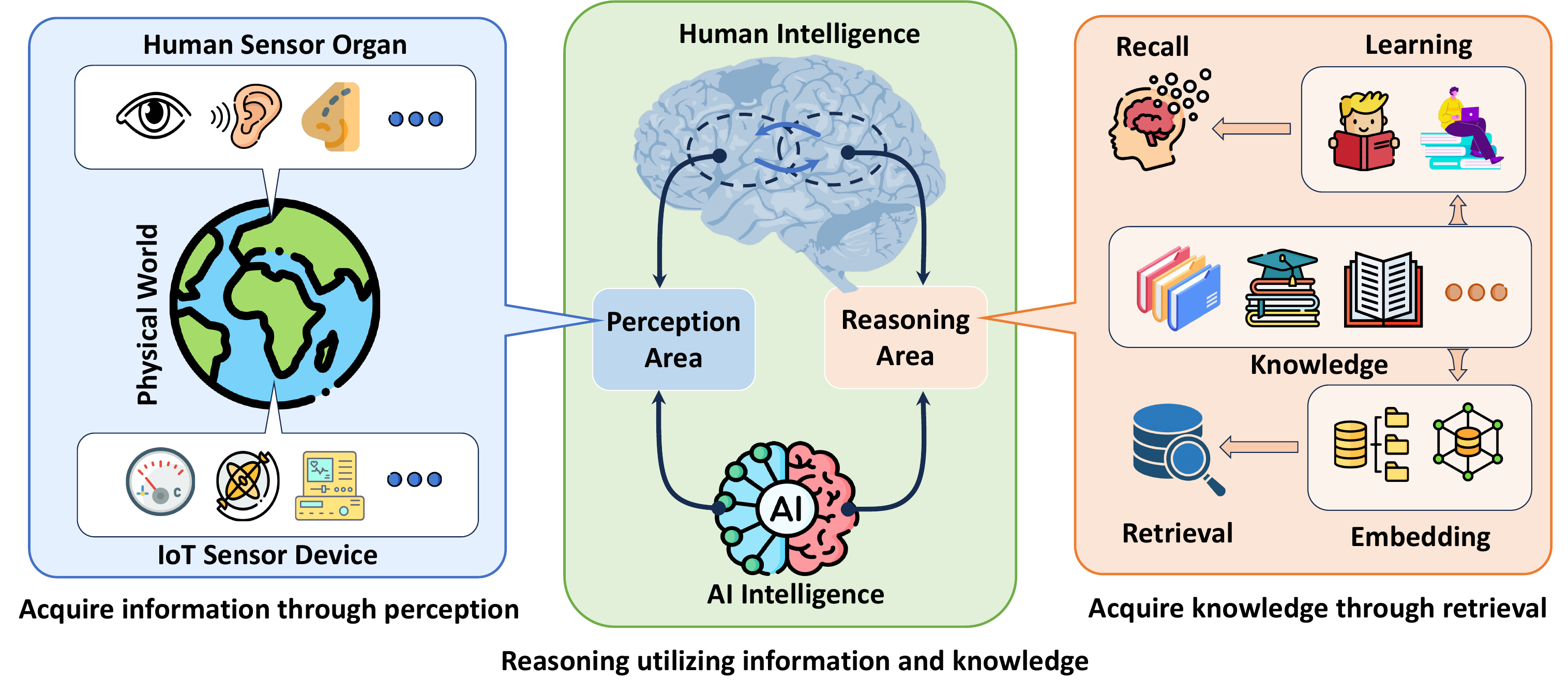}
\end{center}
\caption{Inspired by human cognitive science, we augment LLMs with physical world perception from IoT data. Furthermore, by retrieving pertinent knowledge about IoT tasks, we enhance the reasoning capabilities of LLMs in executing real-world applications.}
\label{fig: figure 1 concept_brain}
\end{figure*}

Unlike LLMs that map descriptions of the physical world to a latent space and perform reasoning by predicting the text sequence according to the probability, research on human cognitive science illustrates a different mechanism. The human brain comprises multiple mutually-functional areas, of which the important components include the temporal and occipital lobes for perception, and the frontal cortex for reasoning~\citep{churchland1988perspectives,saxe2009brain,hobeika2016general,grezes2001does}.
Notably, perception is the primary mechanism through which information about the physical world is acquired, and then effective reasoning is inherently dependent on accurate perception. However, in LLMs, the physical world is only ``perceived'' through natural language, i.e., concepts and words in the semantic space, which denotes an indirect representation and abstraction of the physical world. A recent study in Nature shows language is primarily a tool for communication rather than thought~\citep{fedorenko2024language}, so reasoning the physical-world problem with only language is limited. To enable LLMs with better reasoning capability in the real world, perception is highly demanded. Recent research on Vison Language Models (VLMs) builds the connection between visual perception and languages~\citep{zhang2024visionlanguage}, yet the vision is only one of the various perceptual modalities. Many aspects of the physical world are still not perceived by existing LLMs.

We draw inspiration from human cognition, which integrates perception and reasoning with domain knowledge. Humans perceive the world through sensory organs (e.g., eyes and ears), while IoT sensors serve as analogous "sensory organs" for machines, capturing physical-world data for automation. Since the first IoT sensor was introduced in the 1980s to monitor Coke machine inventories~\citep{madakam2015internet}, these sensors have become essential for modeling the physical environment. 
Humans further process perception data using domain knowledge acquired through experience and education. Similarly, large language models (LLMs) can leverage in-context learning to integrate domain knowledge about the physical world and IoT sensors, enhancing their reasoning capabilities. As illustrated in Fig.~\ref{fig: figure 1 concept_brain}, we hypothesize that combining IoT-enabled perception data with relevant knowledge can enable LLMs to tackle complex real-world tasks. 
In this work, we aim to explore three key questions: (1) What types of real-world tasks can LLMs perform using IoT-enabled perception? (2) How can LLM capabilities be further enhanced for such tasks? (3) Do LLMs truly comprehend perception data and apply knowledge effectively to solve real-world problems?

\begin{figure*}[t!]
\begin{center}
\includegraphics[width=1.0\textwidth]{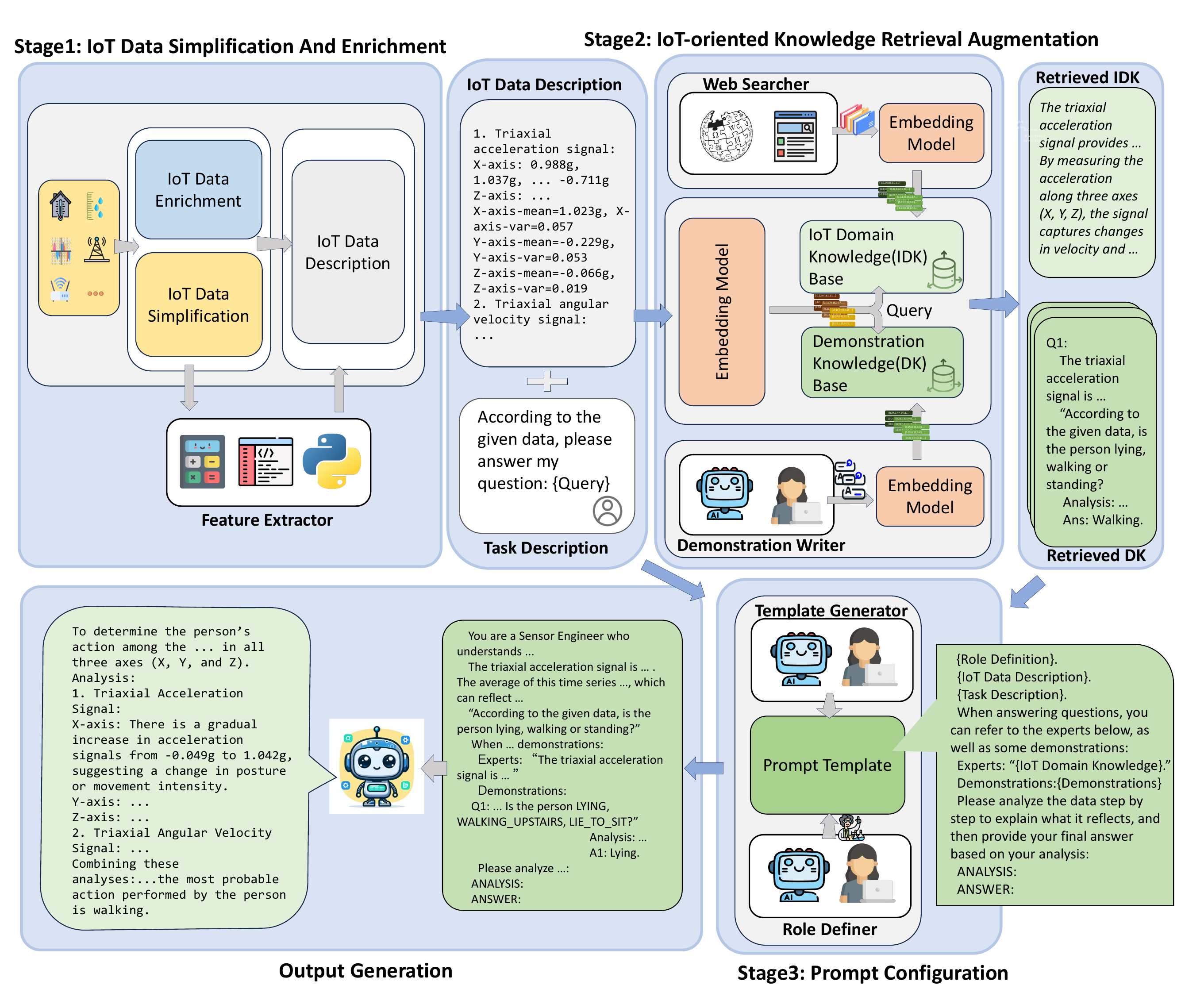}
\end{center}
\caption{In our framework, IoT data is initially preprocessed to create a data description. Next, relevant IoT domain knowledge and task-specific demonstrations are retrieved. These elements are then combined into a prompt, which is input into a LLM to generate the final output.}
\label{fig: figure 2 overall pipeline}
\end{figure*}

Previous studies have primarily shown the viability of using LLMs for IoT-sensory task reasoning~\citep{xu2024penetrative,ji2024hargpt}, but we find that these studies are not carefully scrutinized. (1) These studies only focus on specific tasks, such as R-peak identification and action recognition. The choices of tasks are not comprehensive, thus lacking a benchmark to evaluate the performance of the methods. (2) They directly input raw IoT data into LLMs for reasoning, but LLMs are not good at dense numerical data and calculation~\citep{zhou2024larger,gruver2024large}.
(3) They only evaluate their effectiveness on closed-source LLMs, and lack a comprehensive study of benchmarking open-source LLMs with different parameter size. (4) While some of them also incorporate external domain knowledge to aid LLMs in task handling, they rely on expert-designed knowledge, which is costly, time-consuming, and challenging to adapt to additional tasks due to the need for bespoke expert input.

To bridge this gap and address the proposed questions, we conduct a comprehensive study on leveraging LLMs for various IoT tasks in the physical world. First, we evaluate LLMs' ability to handle IoT classification and regression challenges by introducing a unified benchmark comprising five representative IoT sensory tasks: human activity recognition, industrial anomaly detection, heartbeat anomaly detection, WiFi-based human sensing, and indoor localization. These tasks encompass a wide range of domains, including daily life, industrial applications, and healthcare, with varying degrees of complexity. The datasets involved cover not only common modalities such as IMU readings and industrial sensor measurements (e.g., temperature, pressure) but also high-dimensional signals such as electrocardiogram (ECG) data and even higher-dimensional inputs like Wi-Fi channel state information (CSI).
Next, we propose IoT-LLM, a unified framework to enhance LLMs' reasoning capabilities with IoT data. IoT-LLM includes three key components: IoT data simplification and 
 enrichment, IoT-oriented knowledge retrieval augmentation, and prompt configuration.
Finally, we assess LLMs' comprehension and problem-solving abilities by analyzing their generated reasoning processes. Results demonstrate that IoT-LLM enables LLMs to perform more precise and domain-specific analyses compared to baseline methods, achieving expert-level insights across IoT tasks. In addition, because the domain knowledge in our framework is automatically retrieved from the knowledge base, there is no need to manually craft expert knowledge for each additional task. When adapting to an additional IoT task, simply adding relevant documents to the IoT knowledge base is sufficient, and the rest of the framework’s pipeline remains unchanged. In real-world scenarios, it would be impractical to individually tailor expert knowledge for thousands of IoT tasks, which underscores the advantage of our approach. In summary, our contributions are as follows:
\begin{itemize}
\item We systematically study how LLMs can address real-world problems by perceiving the physical world via IoT sensor data.

\item We propose a unified framework to address IoT-related real-world problems, which enhances the capability of LLMs through three tailored steps. To the best of our knowledge, this is the first unified framework for IoT-sensory tasks in the physical world.

\item We establish the first benchmark for IoT-sensory task reasoning, including five real-world tasks with various types of IoT data. Empirical results show that our IoT-LLM significantly improves the performances of all base LLMs on IoT-sensory tasks. 
\end{itemize}

\section*{RESULTS}


\subsection*{IoT-LLM framework overview}
To begin, we formally define the problem of IoT sensory task reasoning using large language models (LLMs). We then introduce the IoT-LLM framework, which is specifically designed to enhance the reasoning capabilities of LLMs when addressing complex IoT sensory tasks.

The formulated research problem is how to leverage LLM and in-context learning for task reasoning for IoT data, termed as \textit{IoT sensory task reasoning}. Typical examples include using accelerometer data for human activity recognition or machine sensor data for anomaly detection. In this context, the LLM prompt consists of two essential components: (1) the data, which serves as a way to perceive the physical world, and (2) the task description, such as ``Is this a normal heartbeat (N) or a premature ventricular contraction (V)?", which acts as the query in a heartbeat anomaly detection task.

\begin{table*}[!h]
\begin{center}
\scalebox{0.8}{
    \begin{tabular}{cc|cccccc}
    \toprule
    \multicolumn{2}{c}{\multirow{2}{*}{\textbf{Method}}} & \multicolumn{6}{|c}{\textbf{Model}} \\
    
    \cmidrule{3-8}
    
     & & Llama2-7B & Mistral-7B & Claude-3.5 & Gemini-pro & GPT-3.5 & GPT-4o-mini\\
    \midrule
    \multirow{3}{*}{\makecell{Base-\\line}} & \makecell{RMSE (m)} & 0.374 & 11.570 & 0.829 & 2.318 & 2.598 & 0.741\\
    \multirow{3}{*}{} & MAE (m) & 0.313 & 9.347 & 0.696 & 1.814 & 1.937 & 0.581\\
    \multirow{3}{*}{} & STD & 0.903 & 6.856 & 1.607 & 5.999 & 6.715 & 1.502\\
    \midrule
    \multirow{3}{*}{Ours} & RMSE (m) & 0.355 & 9.995 & 0.404 & \textbf{0.313} & 0.719 & 0.402 \\
    \multirow{3}{*}{} & MAE (m) &  0.295 & 7.980 & 0.341 &\textbf{ 0.265} & 0.592 & 0.341 \\
    \multirow{3}{*}{} & STD &  0.852 & 11.146 & \textbf{0.706 }& 0.763 & 1.765 & 0.697 \\
    \midrule
    \multirow{2}{*}{\makecell{Impro-\\vement}} & RMSE (m) &  \textbf{{+5.1\%}} &  \textbf{{+13.6\%}} &  \textbf{{+51.3\%}} &  \textbf{{+86.5\%}} &  \textbf{{+72.3\%}} &  \textbf{{+45.7\%}}\\
    & MAE (m)  &  \textbf{{+5.8\%}} &  \textbf{{+14.6\%}} &  \textbf{{+51.0\%}} &  \textbf{{+85.4\%}} &  \textbf{{+69.4\%}} &  \textbf{{+41.3\%}}\\ 
    \bottomrule
    \end{tabular}
}
\end{center}
\caption{\textbf{Performance of LLMs on WiFi-based Indoor Localization task.} Since this is a regression task, we choose the Root Mean Square Error (RMSE), Mean Absolute Error (MAE), and standard deviation (STD) of the RMSE as the main performance metrics.}
\label{local}

\begin{center}
\scalebox{0.8}{
    \begin{tabular}{c|c|ccccc}
    \toprule
    \multicolumn{2}{c|}{\multirow{2}{*}{\textbf{Model}}} &\multicolumn{5}{c
    }{\textbf{IoT tasks} (Accuracy~$\uparrow$)}\\
    \cmidrule{3-7}
    \multicolumn{2}{c|}{} & \textbf{HAR-2cls} & \textbf{HAR-3cls} &\textbf{Heartbeat} & \textbf{Machine} & \textbf{Occupancy}\\
    \midrule 
    \multirow{3}{*}{Llama2-7B} & Baseline & 50.0\% & 32.8\% & 50.0\% & 35.0\%& 48.4\%\\
    \cmidrule{2-7}
    \multirow{3}{*}{} & Ours & 57.2\% & 38.0\% & 54.5\% &56.4\%&82.5\%\\
    \cmidrule{2-7}
    \multirow{3}{*}{} & Improvement & \textbf{{+14.4\%}} &  \textbf{{+15.9\%}} &   \textbf{{+9.0\%}} &   \textbf{{+61.1\%}} &  \textbf{{+70.5\%}}\\
    \midrule
    \multirow{3}{*}{Mistral-7B} & Baseline & 66.0\% & 21.8\% & 46.0\% & 50.0\%& 50.0\%\\
    \cmidrule{2-7}
    \multirow{3}{*}{} & Ours & 89.0\% & 54.0\% & 66.0\% &\underline{92.1}\%&61.1\%\\
    \cmidrule{2-7}
    \multirow{3}{*}{} &  Improvement &   \textbf{{+34.8\%}} &   \textbf{{+147.7\%}} &   \textbf{{+43.5\%}} &   \textbf{{+84.2\%}} &   \textbf{{+22.2\%}}\\
    \midrule 
    \multirow{3}{*}{Claude-3.5} & Baseline & 98.4\% & 72.0\% & 52.0\% & 50.5\%& 50.0\%\\
    \cmidrule{2-7}
    \multirow{3}{*}{} & Ours & \textbf{100.0}\% & \textbf{98.7}\% & \textbf{83.0}\% &86.3\%&82.5\%\\
    \cmidrule{2-7}
    \multirow{3}{*}{} &  Improvement &   \textbf{{+1.6\%}} &   \textbf{{+37.1\%}} &   \textbf{{+59.6\%}} &   \textbf{{+70.9\%}} &   \textbf{{+65.0\%}}\\
    \midrule
    \multirow{3}{*}{Gemini-pro} & Baseline & 96.0\% & 56.7\% & 49.0\% & 48.5\%& 55.9\%\\
    \cmidrule{2-7}
    \multirow{3}{*}{} & Ours & \underline{98.0}\% & \underline{82.8}\% & \underline{73.5}\% &70.1\%&66.2\%\\
    \cmidrule{2-7}
    \multirow{3}{*}{} &  Improvement &   \textbf{{+2.1\%}} &   \textbf{{+46.0\%}} &   \textbf{{+50.0\%}} &  \textbf{{+44.5\%}} &  \textbf{{+18.4\%}}\\
    \midrule
    \multirow{3}{*}{GPT-3.5} & Baseline & 81.0\% & 40.7\% & 37.0\% & 49.5\%& 50.0\%\\
    \cmidrule{2-7}
    \multirow{2}{*}{} & Ours & 92.1\% & 55.3\% & 58.5\% &61.5\%&\underline{92.1}\%\\
    \cmidrule{2-7}
    \multirow{3}{*}{} &  Improvement &  \textbf{{+13.7\%}} &  \textbf{{+35.9\%}} &  \textbf{{+58.1\%}} &  \textbf{{+24.2\%}} &  \textbf{{+84.2\%}}\\
    \midrule 
    \multirow{3}{*}{GPT-4o-mini} & Baseline & 89.0\% & 37.3\% & 44.0\% & 57.0\%& 82.3\%\\
    \cmidrule{2-7}
    \multirow{3}{*}{} & Ours & \textbf{100.0}\% & 77.8\% & 68.0\% &\textbf{92.5}\%&\textbf{92.7}\%\\
    \cmidrule{2-7}
    \multirow{3}{*}{} &  Improvement & \textbf{{+12.4\%}} &  \textbf{{+108.6\%}} &  \textbf{{+54.5\%}} &  \textbf{{+62.3\%}} &  \textbf{{+12.6\%}}\\
    \bottomrule
    \end{tabular}
}
\caption{\textbf{Average performance of LLMs on IoT tasks}. \textbf{HAR-2cls} stands for classifying walking and standing activities. \textbf{HAR-3cls} stands for classifying lying, walking upstairs, and transitioning from lying to sitting activities. \textbf{Heartbeat} stands for classifying normal and abnormal heartbeats. \textbf{Machine} stands for determining whether the coolers work properly or not. \textbf{Occupancy} stands for detecting the presence of a person in a room.}
\label{main results}
\end{center}
\end{table*}

To systematically assess the ability of current LLMs to handle IoT sensory tasks, we construct a unified benchmark comprising five real-world tasks that span diverse IoT data types and varying levels of difficulty, covering both classification and regression problems. Initially, we evaluate LLMs in a basic setting, akin to Penetrative AI~\citep{xu2024penetrative}, where the prompt provided to the LLM includes raw IoT data, the associated query, and optionally, manually curated expert knowledge. However, the performance of LLMs in this setting remains suboptimal. As shown by the baseline results in Table~\ref{main results}, even GPT-4o-mini achieves only 37.3\% accuracy for 3-way activity recognition and 44\% for heartbeat anomaly detection. These results, which are close to random guessing, indicate that naive prompting fails to enable LLMs to comprehend IoT data and tasks effectively.

Through an analysis of IoT data characteristics and IoT sensory task requirements, we identify two primary challenges: the abstraction of dense numerical IoT data and the lack of domain-specific knowledge within LLMs. To address these issues, we propose a unified framework (Fig.~\ref{fig: figure 2 overall pipeline}) comprising three key stages: (1) IoT data simplification and enrichment, (2) IoT-oriented knowledge augmentation, and (3) prompt configuration. Each stage is designed to tackle specific obstacles faced by LLMs in IoT sensory task reasoning.

In the IoT data simplification and enrichment stage, we focus on enhancing the physical interpretability of the data and simplifying complex, long-sequence raw IoT data. This processing makes it easier for LLMs to understand the data and its underlying physical meaning, thereby improving their ability to utilize the information for IoT sensory task reasoning.

In the IoT-oriented knowledge augmentation stage, we address the challenge that LLMs often lack the specialized domain knowledge required for certain tasks. We automatically retrieve relevant domain knowledge from constructed knowledge bases, providing LLMs with expert-level information through in-context learning. Notably, unlike prior approaches that rely on manually curated expert knowledge for each specific task~\citep{xu2024penetrative, ji2024hargpt}, our automated retrieval method eliminates the need for manual annotation, significantly reducing human effort and making it more practical to scale to additional IoT tasks.

Finally, in the prompt configuration stage, all LLM-friendly IoT data descriptions generated in the previous stages, along with the retrieved domain documents, are integrated into a predefined prompt template. This results in the final prompt, which is then input to the downstream LLM for task execution.

\subsection*{IoT Reasoning Benchmark}
\paragraph{IoT Sensory Tasks.}
To comprehensively assess the capability boundaries of LLMs for IoT-sensory task reasoning, we develop a unified benchmark comprising five real-world tasks with diverse IoT data types and difficulty levels: (1) Human Activity Recognition (HAR) using Inertial Measurement Unit (IMU) data, (2) Industrial anomaly detection using metrics such as temperature, cooling power, and cooling efficiency, (3) Heartbeat anomaly detection using Electrocardiogram (ECG) data, (4) Human sensing using WiFi Channel State Information (CSI), and (5) Indoor localization based on WiFi signal strength.

\paragraph{LLM baselines.} In the conducted experiments, we utilize a combination of proprietary and open-source LLMs, including GPT-3.5-turbo, GPT-4o-mini, claude-3.5-sonnet, gemini-pro, \href{https://huggingface.co/mistralai/Mistral-7B-Instruct-v0.3}{Mistral-7B}, and \href{https://huggingface.co/togethercomputer/LLaMA-2-7B-32K}{LLama2-7B}. This diverse selection of models enables a comprehensive evaluation of the LLMs' capabilities in executing IoT-sensory tasks and provides insights into their respective strengths and limitations in real-world applications. 

\paragraph{Baseline method.}
To evaluate the efficacy of our proposed framework in enhancing the capabilities of IoT-sensory task reasoning for LLMs, we employ Penetrative AI~\citep{xu2024penetrative} as a baseline method. This involves utilizing prompts that incorporate raw IoT data, corresponding task descriptions, and, optionally, specially designed expert knowledge. To achieve this, we strictly adhere to the methodology outlined in the Penetrative AI paper, ensuring a rigorous replication of their approach. The overall performance of LLMs on IoT-sensory tasks is shown in Table~\ref{local} and Table~\ref{main results}. The results indicate that our proposed framework significantly enhances the performance of all the LLMs in completing IoT-sensory tasks in real-world scenarios. After analyzing the overall performance of LLMs in the task reasoning, we can answer the questions we proposed in the introduction now. Here is a summary of our arguments regarding the IoT-sensory task reasoning with LLMs.

\begin{figure*}[t!]
\begin{center}
\includegraphics[width=1.0\textwidth]{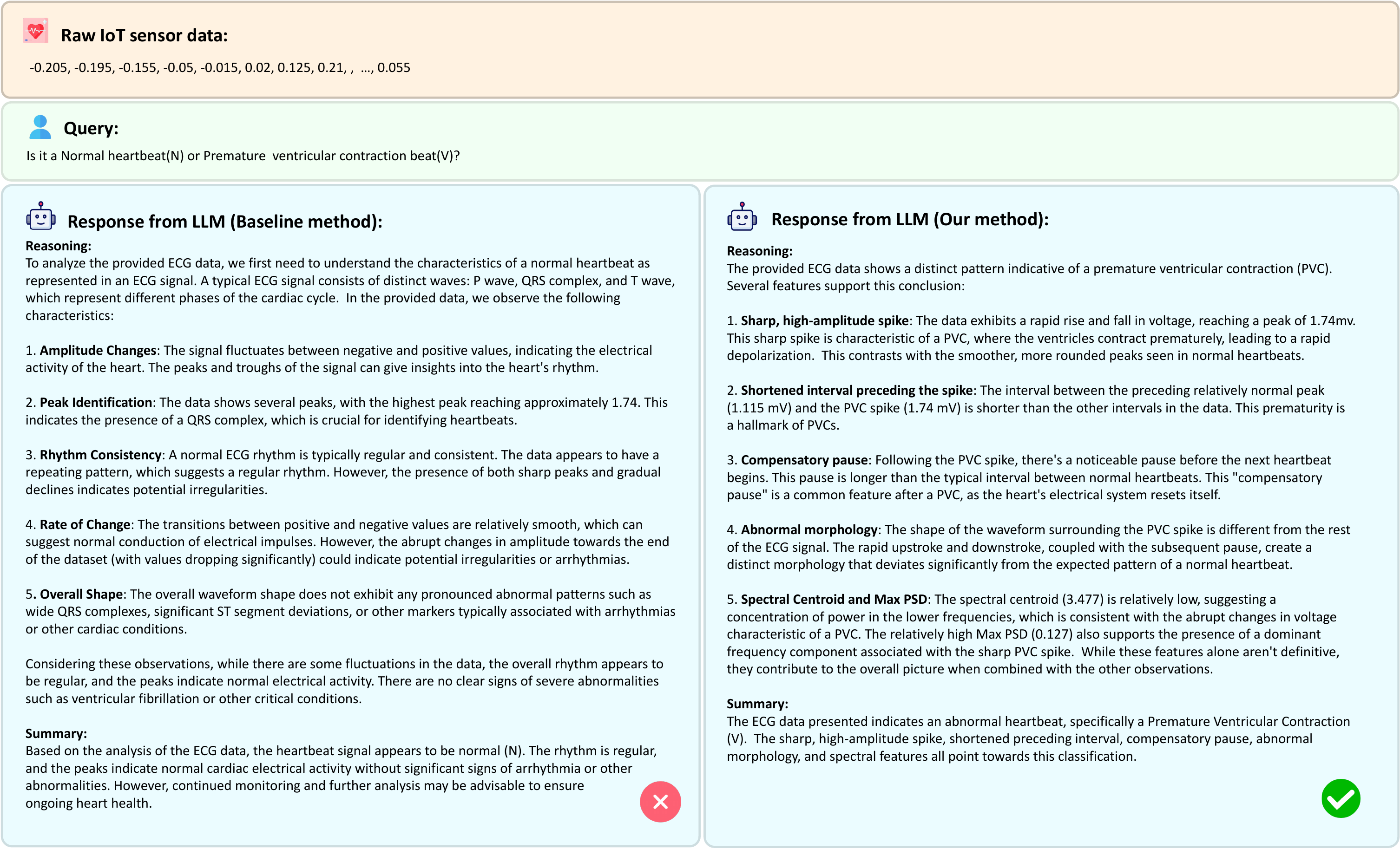}
\end{center}
\caption{\textbf{Response examples comparing the baseline method and our approach in heartbeat anomaly detection}. The baseline method offers logically coherent but generalized analyses, whereas our method provides deeper insights and more precise descriptions of ventricular premature contraction characteristics, resulting in more professional and accurate responses.}
\label{fig: figure 3 ecg compare}
\end{figure*}

\subsection*{LLMs excel in various IoT-sensory tasks but struggle with complex data challenge.}
Experimental results indicate that advanced LLMs, such as GPT-4o-mini and Claude-3.5, can effectively perform various IoT-sensory tasks, particularly excelling in HAR using IMU data. However, their performance is limited when faced with more challenging tasks that involve complex data and require specialized knowledge. For example, in heartbeat anomaly detection, LLMs perform sub-optimally due to the time-series nature and numerical complexity of ECG data. Simplifying the data mitigates some challenges but does not address the fundamental model limitations. Additionally, LLMs lack the extensive medical knowledge necessary for comprehensive analysis. While retrieved knowledge may suffice for simpler tasks, more complex problems might require further model fine-tuning to incorporate deeper medical expertise.

\subsection*{LLMs are excellent learners in IoT-sensory task reasoning.} 
Without domain-specific knowledge and relevant demonstrations, LLMs face significant challenges in performing IoT-sensory tasks, often resorting to near-random guessing, especially in tasks such as heartbeat anomaly detection. This indicates that real-world tasks remain challenging for LLMs to execute directly. However, LLMs are excellent learners, and their capabilities can be significantly enhanced through our proposed framework. Specifically, the LLama2-7B, Mistral-7B, Claude-3.5, Gemini-pro, GPT-3.5, and GPT-4o-mini models exhibit average performance improvements of 30\%, 58\%, 48\%, 41\%, 48\%, and 49\% respectively across various tasks, underscoring the effectiveness of our methodology.
\begin{table*}[t!]
\begin{center}
\scalebox{0.85}{
    \begin{tabular}{c|cccccc}
    \toprule
    \multirow{2}{*}{ \textbf{Method}} & \multicolumn{5}{c}{\textbf{IoT tasks}~(Accuracy~$\uparrow$)}\\
    \cmidrule{2-6}
    \multirow{2}{*}{} & \textbf{HAR-2cls} & \textbf{HAR-3cls} & \textbf{Machine} & \textbf{Heartbeat} & \textbf{Occupancy} \\
    \midrule
    Baseline & 89.0\% & 37.3\% & 57.0\% & 44.0\% & 82.3\% & \\
    \midrule
    +~\textit{IoT data simplification and enrichment} & 92.0\% & 43.3\% & 58.3\% & 53.0\% & 83.0\%\\
    \midrule
    +~\textit{retrieved domain knowledge} & 92.4\% & 59.3\% & 66.4\% & 55.3\% & 85.3\%\\
    \midrule
    +~\textit{retrieved demonstrations} & 99.0\% & 66.7\% & 86.4\% & 64.0\% & 87.7\%\\
    \midrule
    Full setting & \textbf{100.0\%} & \textbf{77.8\%} & \textbf{92.5\% } & \textbf{68.0\%} & \textbf{92.7\%}\\
    \bottomrule
    \end{tabular}
}
\end{center}
\caption{Ablation study of different modules within our framework.}
\label{ablation}
\end{table*}

\subsection*{LLMs can address tasks from an expert perspective.}
Compared with the baseline method, our framework enables LLMs to perform in-depth, targeted data analysis akin to that of a domain expert. While the baseline allows LLMs to provide logically consistent but often superficial analyses based on IoT data—lacking specificity and deep exploration of data characteristics—our framework enhances their capabilities through advanced data and knowledge integration. This results in more comprehensive analyses and professional responses.
For example, in heartbeat anomaly detection (see Fig.~\ref{fig: figure 3 ecg compare}, more examples can be seen in Fig.~\ref{fig: HAR} - Fig.~\ref{fig: localization} in the appendix), the baseline method mentions general features like peaks and rhythms but offers only broad descriptions of fluctuations, failing to identify critical abnormalities such as compensatory pauses and waveform irregularities. This leads to vague and incomplete reasoning. In contrast, our method meticulously identifies and examines key features, such as premature ventricular contractions (PVCs) characterized by early ventricular contractions followed by compensatory pauses, which are essential for accurate PVC diagnosis.
This comparison demonstrates that by enhancing perceptual and knowledge-based capabilities, our framework significantly improves LLMs' ability to deeply understand IoT data and its physical context, enabling precise, expert-level analysis.

\subsection*{Ablation study and retrieval sensitivity analysis}
\subsubsection*{Ablation study}
To assess the contribution of each component in \emph{IoT-LLM}, we conduct an ablation using GPT-4o-mini under four incremental configurations: (1) \emph{IoT data simplification \& enrichment}; (2) + \emph{retrieved domain knowledge}; (3) + \emph{retrieved demonstrations}; and (4) the \emph{full} setting that further incorporates \emph{role prompting} and \emph{chain-of-thought} as specified in the Prompt Configuration stage. To isolate module effects and avoid confounding, we use a validated retrieval setting throughout; sensitivity to retrieval parameters will be analyzed separately in the following section.

The results presented in Table~\ref{ablation} show a consistent, step-wise improvement. For relatively straightforward tasks (e.g., HAR-2cls), \emph{IoT data simplification \& enrichment} already helps by stabilizing tokenization and exposing salient statistics, and adding \emph{retrieved domain knowledge} is sufficient to reach strong performance. For more challenging settings (e.g., HAR-3cls, Machine anomaly, Heartbeat, and Occupancy), \emph{retrieved demonstrations} further guide label mapping and decision heuristics, and the \emph{full} configuration yields the best overall accuracy by eliciting multi-step reasoning.

In summary, the ablation shows that the components work well together and add up. Data simplification \& enrichment improves numeric stability and highlights key patterns in the signal; retrieved domain knowledge provides task-specific context; retrieved demonstrations clarify label meanings and typical decision rules; and the final prompting (role prompting with chain-of-thought) guides step-by-step reasoning. Used together, these parts deliver a clear, steady improvement over partial settings and help the LLM reason more reliably on IoT-sensory tasks.

\begin{figure*}[t!]
\begin{center}
\includegraphics[width=1.0\textwidth]{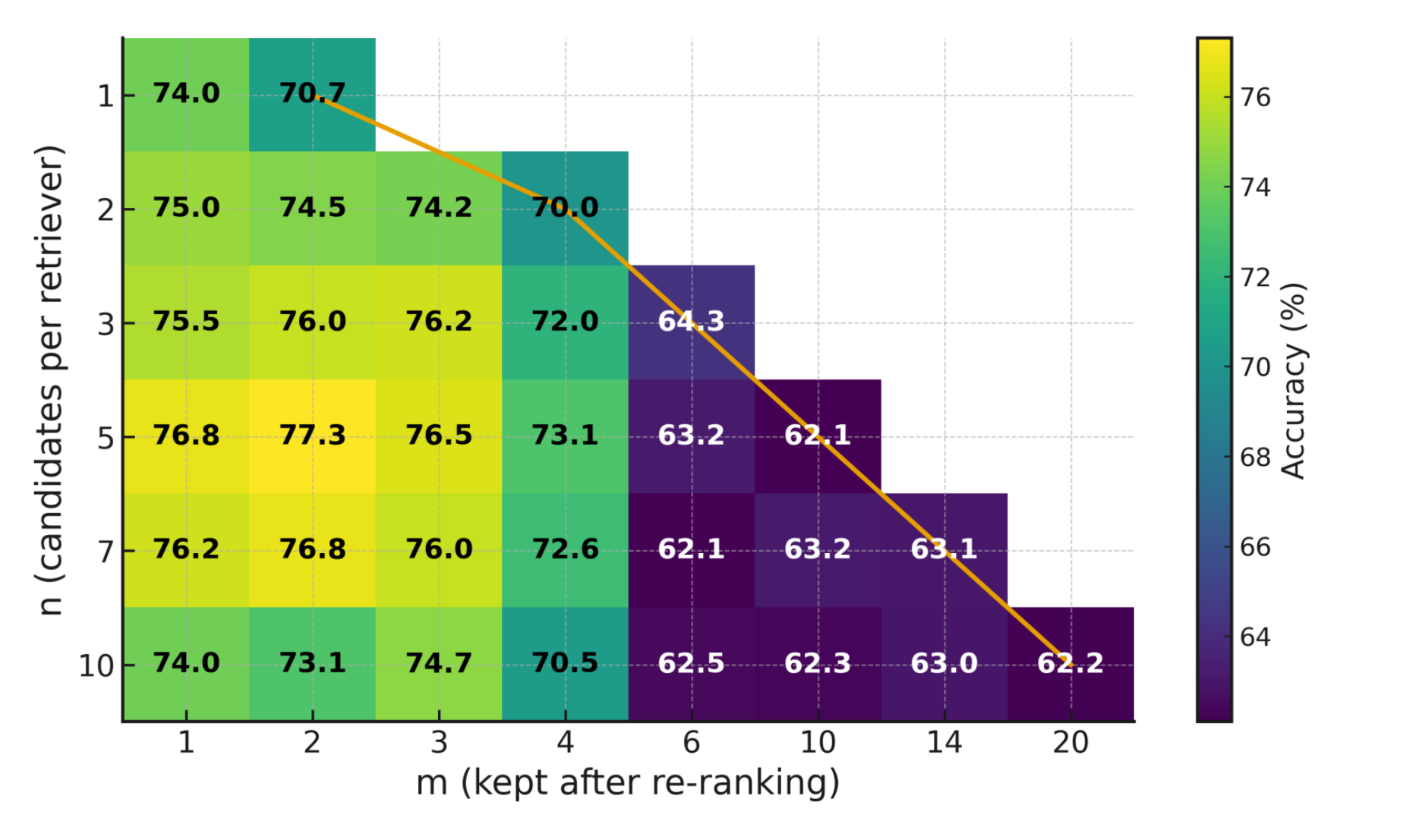}
\end{center}
\caption{\textbf{Accuracy heatmap over $(n,m)$.} Cells report accuracy (\%) for candidates per retriever $n$ (rows) and kept passages after re-ranking $m$ (columns).
Missing cells are \emph{infeasible} ($m>2n$); the orange polyline marks the feasibility boundary $m{=}2n$.
The optimum appears at a small keep set ($m{=}2$) with moderate candidates ($n{=}5$); accuracy degrades for larger $m$, while increasing $n$ beyond $\sim\!5$ yields diminishing returns.}
\label{fig: figure 7 heatmap}
\end{figure*}

\subsubsection*{Retrieval sensitivity analysis}
\begin{figure*}[t!]
\begin{center}
\includegraphics[width=1.0\textwidth]{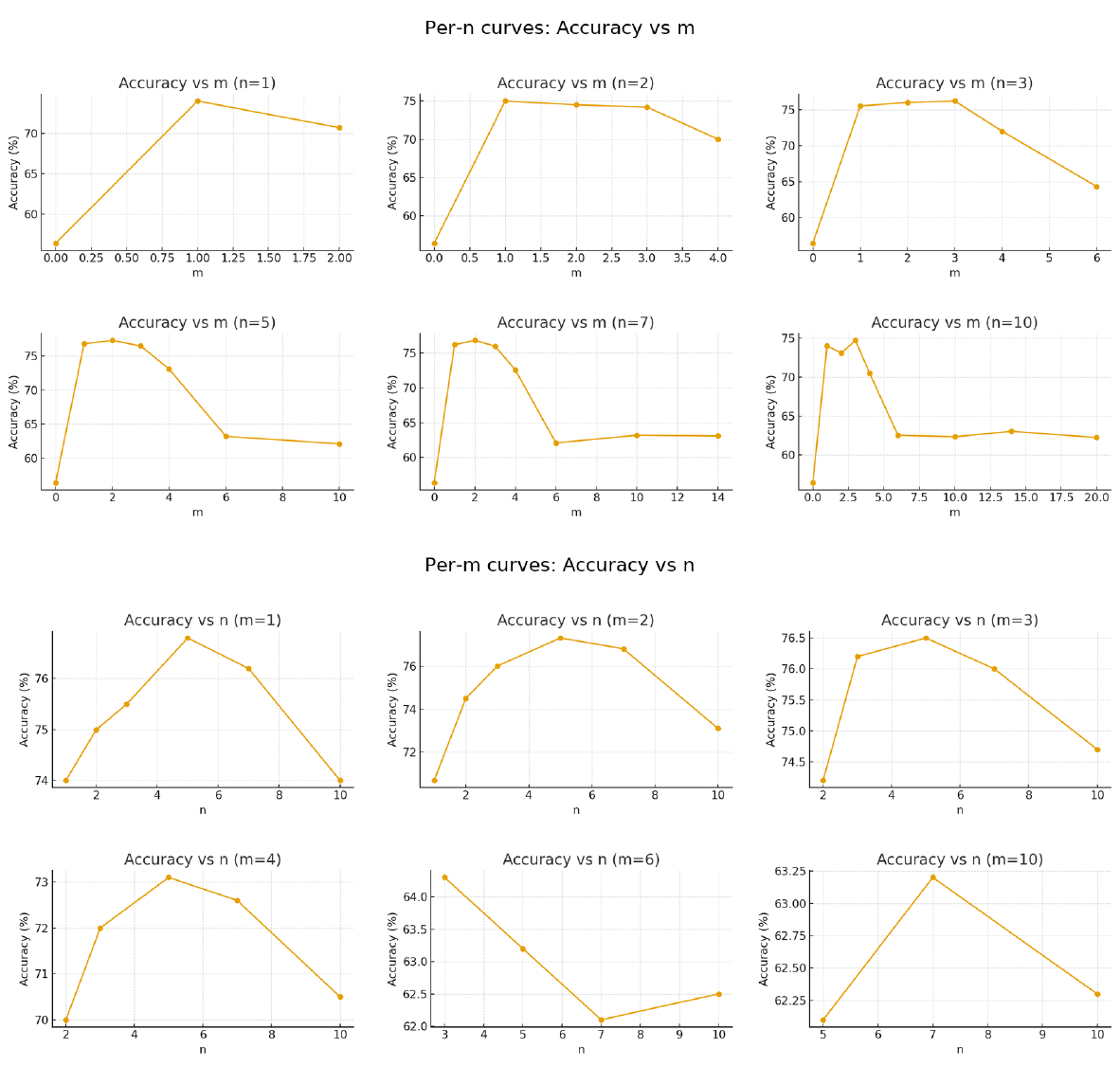}
\end{center}
\caption{\textbf{Retrieval quantity sensitivity—per-$n$ and per-$m$ views.}
\emph{Top:} Accuracy vs.\ kept passages $m$ for fixed $n\in\{1,2,3,5,7,10\}$.
\emph{Bottom:} Accuracy vs.\ candidates $n$ for fixed $m\in\{1,2,3,4,6,10\}$.
Across settings, accuracy rises steeply from $m{=}0$ to a small keep set and peaks at $m\in\{2,3\}$, then declines as $m$ grows, consistent with noise/attention dilution.
Increasing $n$ helps up to a saturation around $n\!\approx\!5$, after which returns diminish.
Together these views show that a \emph{small $m$} and a \emph{moderate $n$} are sufficient and robust.}
\label{fig: figure 8 mn_sensitivity}
\end{figure*}

During retrieval, we employ two retriever types (sparse and dense) to fetch 2n candidate paragraphs (n paragraphs per retriever), followed by a cross-encoder re-ranker that selects the top-m passages based on the similarity between the retrieved content and the query to serve as domain knowledge to feed the LLM (details in Method section). To assess sensitivity to retrieval depth and keep ratio, we vary $n\!\in\!\{1,2,3,5,7,10\}$ and $m\!\in\!\{0,1,2,3,4,6,10,14,20\}$ (subject to $m\!\le\!2n$) and evaluate the 3-class HAR task with GPT-4o-mini. The accuracy heatmap is shown in Fig.\ref{fig: figure 7 heatmap}, while the corresponding m/n sensitivity is provided in Fig.~\ref{fig: figure 8 mn_sensitivity}. From these results, we draw the following insights.

\textbf{Overall gains and saturation.} Even a small amount of retrieved knowledge yields a large improvement over the no-retrieval domain knowledge baseline ($m\!=\!0$: $56.4\%$). With only a few passages, accuracy rises to the $74\%\!\sim\!77\%$ range. The best result is attained at a \emph{moderate} candidate pool and a \emph{small} keep ratio, peaking at $(n,m)\!=\!(5,2)$ with \textbf{77.3\%}. A broad near-optimal plateau is observed around $n\!\in\!\{3,5,7\}$ and $m\!\in\!\{1,2,3\}$---e.g., $(5,1)$: $76.8\%$, $(5,3)$: $76.5\%$, $(3,3)$: $76.2\%$, $(7,2)$: $76.8\%$. Beyond this region, increasing $m$ (and thus context length) leads to diminishing or negative returns.

\textbf{Too many passages introduce noise.} For a fixed $n$, accuracy typically degrades as $m$ grows: at $n\!=\!5$, performance drops from $77.3\%$ ($m\!=\!2$) to $73.1\%$ ($m\!=\!4$), $63.2\%$ ($m\!=\!6$) and $62.1\%$ ($m\!=\!10$). A similar trend holds for $n\!=\!7$ and $n\!=\!10$. This suggests that (i) longer prompts accumulate redundant or off-topic paragraphs that dilute salient evidence, and (ii) re-ranking, while helpful, cannot fully suppress topical drift when the keep set becomes large.

\textbf{Too small a pool can also hurt.} When $n$ is very small, adding more kept passages can reduce accuracy because the re-ranker has limited headroom to separate signal from noise (e.g., at $n\!=\!1$, $m\!=\!1$: $74.0\%$ vs.\ $m\!=\!2$: $70.7\%$). This highlights the need for a minimally sufficient candidate pool \emph{and} a conservative keep ratio.

In summary, retrieval brings large benefits with only a handful of passages, and performance \emph{saturates}---or even degrades---when the kept set grows beyond a small, high-precision subset. These findings justify our design choice of hybrid retrieval plus re-ranking, and they provide actionable defaults for both accuracy-oriented and edge-oriented settings.

\section*{DISCUSSION}

\subsection*{Summary}
Large Language Models (LLMs) have demonstrated remarkable performance in traditional domains such as natural language processing and computer vision. However, these tasks are primarily confined to the digital realm. In contrast, LLMs often struggle with real-world tasks that require interaction with the physical environment. Meanwhile, the proliferation of IoT devices has enabled the collection of diverse and rich data from the physical world, which can be leveraged to address a wide range of practical tasks. This raises an important question: how can we effectively utilize IoT data to solve real-world problems within a unified framework?
Given that LLMs are pre-trained on large-scale datasets and possess extensive common-sense knowledge, we hypothesize that they are well-suited to serve as the core model in such a framework. To further enhance the ability of LLMs to process and reason over IoT sensory data, we propose a unified framework, IoT-LLM, which aims to improve both the perception and reasoning capabilities of LLMs.

To comprehensively evaluate the effectiveness of LLMs on IoT-sensory tasks, we establish the first benchmark for IoT-sensory task reasoning, encompassing five real-world tasks with diverse types of IoT data. In our benchmark, we compare our method with a baseline approach. The results clearly demonstrate that our method significantly outperforms the baseline across all LLMs and IoT tasks. For relatively simple tasks, such as human activity recognition, our approach achieves substantial performance improvements. For more challenging tasks, such as heartbeat anomaly detection, the accuracy of LLMs increases from near-random guessing to 83\%. These findings validate the feasibility of our framework and demonstrate that leveraging LLMs with IoT data can effectively address real-world physical tasks.

Beyond quantitative metrics, LLMs also provide deeper analytical insights compared to baseline methods. For instance, in the heartbeat anomaly detection task, LLMs can emulate the reasoning process of a professional diagnostician, offering well-founded inferences based on features such as the QRS complex, T wave, and P wave. This level of interpretability and persuasiveness stands in stark contrast to end-to-end deep learning models trained solely for specific domain tasks, which typically only output final predictions. Our results further confirm that LLMs have the potential to utilize provided knowledge to perform robust reasoning on complex tasks.

Last but not least, it is important to highlight that, unlike previous studies which primarily focus on deploying LLMs within specific domains, the proposed IoT-LLM framework provides a generalizable and unified approach applicable to a wide range of IoT-sensory tasks. This framework can be readily extended to additional tasks, provided that relevant IoT data is available. In this study, we design and implement five representative IoT-sensory tasks in real-world settings, encompassing data modalities ranging from conventional IMU measurements to high-dimensional WiFi CSI signals. The pronounced performance improvements achieved across these tasks provide strong empirical evidence for the effectiveness and superiority of the framework. The unified nature of our approach facilitates easy adaptation to additional scenarios. We believe that, in the future, LLMs will not only serve as valuable assistants for digital tasks such as text polishing, but will also have a broader and more profound impact on real-world physical tasks, thereby offering significant benefits in various aspects of everyday life.

\subsection*{Challenges and future directions}

In our experiments, the proposed IoT–LLM framework has been deployed across a diverse range of IoT sensory tasks. The data types span from low-dimensional signals such as IMU measurements to higher-dimensional inputs such as WiFi Channel State Information (CSI) data. For example, our WiFi CSI dataset has the shape (T, 3, 114), where T denotes the time-series length, 3 represents the number of antennas, and 114 corresponds to the subcarrier dimension for each antenna. This inherently constitutes a high-dimensional temporal dataset, and in our implementation, we applied feature compression and dimensionality reduction techniques to enable efficient integration with the LLM module.

Nevertheless, the core reasoning component in our framework is a LLM, which is primarily trained on natural language corpora. The capability of such LLMs to directly process super-high-dimensional and cross-modal sensory data (e.g., high-resolution video streams, LiDAR point clouds, or dense 3D reconstructions) remains limited without additional modality adaptation. Such modalities require specialized encoders and tokenization schemes to bridge the gap between raw high-dimensional signals and the textual/embedding space where LLMs excel. For real-world scenarios involving extremely high-dimensional data, we identify two general solution pathways.

\textbf{Feature-based compression for text-convertible modalities.}
For modalities whose data structure can be naturally transformed into sequential or symbolic representations—such as high-dimensional time-series (e.g., WiFi CSI, millimeter-wave radar, EEG, ECG) or structured event logs—it is effective to apply feature selection or dimensionality reduction to extract salient information before passing it to the LLM. For example, classical dimensionality reduction techniques such as Principal Component Analysis (PCA), Linear Discriminant Analysis (LDA), t-distributed Stochastic Neighbor Embedding (t-SNE), and Uniform Manifold Approximation and Projection (UMAP) can be employed to extract the most informative components from high-dimensional complex data while reducing redundancy\citep{abdi2010principal, tharwat2017linear, cieslak2020t, mcinnes2018umap}. In our human sensing experiments with WiFi CSI data, we adopted a PCA-based reduction workflow before feeding the processed features into the LLM.

\textbf{Modality-specific encoding for non-textual modalities.}
For sensory modalities that cannot be easily expressed in textual form—such as high-resolution video, raw images, LiDAR point clouds, or unprocessed audio—specialized neural encoders (e.g., CNNs or vision transformers for imagery and video\citep{howard2019searching,tan2019efficientnet,liu2021swin,dosovitskiy2021image}, sparse point cloud encoders for 3D data\citep{qi2017pointnet++,hu2020randla}) are required to transform the raw data into semantic embeddings. These embeddings must then be aligned with the textual embedding space of the LLM, typically via multi-modal pretraining on large-scale paired datasets. This process yields multi-modal large language models (MMLLMs) capable of reasoning over both textual and non-textual inputs. Representative approaches include OneLLM\citep{han2024onellm} and Meta-Transformer\citep{zhang2023meta}, which integrate modality-aware tokenization and embedding alignment to enable unified multi-modal understanding.


In summary, while our current implementation can already process certain high-dimensional data types such as WiFi CSI, extending the IoT–LLM framework to ultra-high-dimensional and cross-modal inputs will require lightweight feature extraction or modality-specific encoders, paired with large-scale multi-modal fine-tuning. Nonetheless, our primary contribution lies in showing that the proposed framework can enhance IoT-specific reasoning capabilities of existing models and in establishing a benchmark to explore LLMs’ capability boundaries on IoT-related real-world tasks. With the continuous advancement of LLMs and their training on increasingly diverse multi-modal datasets, we expect the framework to naturally expand to more complex domains, further improving the generalizability and applicability of LLM-driven IoT systems.

\newpage


\section*{METHODS}


\subsection*{Related works}
Recently, with the rapid advancement of Large Language Models (LLMs), exemplified by the GPT series, there has been a growing interest in leveraging these models to address various challenges within the Internet of Things (IoT) domain. The existing literature primarily explores the application of LLMs in IoT as user interfaces, coordinators among different modules in smart devices, or as third-party assistant tools—such as data generators and analyzers~\citep{li2023chatiot, cui2023llmind, du2023space, yang2025fuzzcode, xia2024fcllm}. For example, LLMind~\citep{cui2023llmind} treats large language models as coordinators and proposes a unified framework that integrates LLMs with AI-specific modules across domains such as IoT, computer vision, and network management. Using LLMs as orchestration agents, this approach enables effective integration of IoT devices and addresses shortcomings in complex task planning and operational efficiency. FuzzCoder~\citep{yang2025fuzzcode} uses LLMs as intelligent mutation generators and optimizers. By fine-tuning on a large corpus of successful mutations (FuzzInstruct), the model learns mutation patterns that trigger crashes or reveal unseen execution paths and predicts optimal mutation positions and strategies in a byte-level sequence-to-sequence framework, delivering a substantial boost to IoT fuzz-testing efficiency and vulnerability discovery.  FCLLM-DT~\citep{xia2024fcllm} uses the LLM as a virtual data generator, leveraging Retrieval-Augmented Generation (RAG) with a knowledge base of historical sensor data to enhance synthetic data quality. During sensor outages, it generates high-quality synthetic data to ensure the continuous operation of IoT-based industrial fault diagnosis and the stability of model training, thereby mitigating accuracy loss caused by data interruptions.

However, in these studies, LLMs primarily serve an auxiliary role rather than directly interpreting IoT data to perform real-world tasks as a core component, which is the central focus of our work. Recent studies, such as Penetrative AI~\citep{xu-etal-2024-penetrative} and HarGPT~\citep{ji2024hargpt}, have begun integrating IoT data into LLMs for specific tasks, leveraging their inherent knowledge bases. Despite these advances, the exploration of LLMs processing IoT data remains nascent. Penetrative AI converts IoT data into textual and numerical formats for basic tasks like R-peak identification in ECG data, heavily relying on manually crafted expert knowledge, which limits automation and scalability. Similarly, HarGPT processes raw IMU data to recognize human activities using a chain of thought technique but is restricted to this specific data type and task, not demonstrating the broader applicability of LLMs. While these studies offer initial insights into using LLMs in the IoT domain, they lack a comprehensive framework to fully leverage LLM capabilities or systematically benchmark current LLMs on IoT sensory tasks, which is the gap our research aims to address.
\subsection*{IoT-LLM framework architecture}

\subsubsection*{IoT data simplification and enrichment}
Unlike textual human tasks that have been learned by LLMs, IoT data for IoT-sensory task reasoning presents unique challenges that hinder LLMs' comprehension. Firstly, IoT data encompasses a diverse range of types and forms, many of which are complex time-series data (e.g., electrocardiogram readings)~\citep{932724} or multi-variant data (e.g., WiFi CSI)~\citep{yang2024mm}. LLMs often struggle with accurately interpreting dense numerical data, especially when it involves long-sequence time-series data~\citep{Zhang2024LargeLM}. Secondly, IoT data is typically composed of raw numerical values. This data often lacks essential textual annotations, such as units of measurement and metadata about the data collection process, which are critical for LLMs to interpret effectively in real-world applications. In summary, raw IoT data requires (1) appropriate simplification and (2) information enrichment. Previous studies~\citep{xu2024penetrative, ji2024hargpt} have employed down-sampling techniques for time-series data but they only achieve coarse-grained simplification at a length level without enhancing the informational content of the IoT data. In contrast, we not only simplify IoT data at the token level but also enrich the IoT data by providing additional information to facilitate better understanding by LLMs. In this way, we transform complex raw IoT data into an LLM-friendly format for IoT-sensory task reasoning (as illustrated in Fig.~\ref{fig: figure 4 data preprocess} in the appendix)

\paragraph{IoT data simplification.}
LLMs face inherent challenges when processing dense numerical streams from IoT sensors, largely due to limitations in tokenization methods such as Byte Pair Encoding (BPE). As shown in prior studies~\citep{gruver2024large, DBLP:journals/jamia/SpathisK24, singh2024tokenization}, numeric values—especially floating-point numbers—are often fragmented into sub-tokens that do not align with their digit structure, leading to inconsistencies and complicating arithmetic reasoning. To address this, we first apply temporal down-sampling and fixed-precision formatting (e.g., two decimal places) to control sequence length, followed by inserting spaces between individual digits to enforce consistent token boundaries, and using commas (",") to separate time steps in a series.

Beyond token-level issues, the inherent complexity and high dimensionality of long-sequence IoT data also make direct interpretation challenging for LLMs. To mitigate this, we extract compact statistical features—such as mean, variance, and FFT mean—using external tools (e.g., Python scripts). These features are provided to the LLM alongside a simplified yet sequential numeric representation, which preserves basic temporal patterns that might be overlooked by purely statistical summaries, while emphasizing salient characteristics to reduce the cognitive load on the model. This design balances information retention with efficiency, leveraging statistical features that have demonstrated strong performance in prior IoT research and proven effective in improving reasoning under context-length constraints\citep{krishnamurthi2020overview, kim2022anomaly, wu2018feature}. In our experiments, this combination of statistical summaries and simplified sequences is sufficient for accurate reasoning in representative IoT sensory tasks. While this approach effectively addresses context limitations, richer temporal dependencies could be modeled in future work using lightweight temporal encoders, symbolic approximations (e.g., SAX\citep{lin2007experiencing}, PAA\citep{keogh2001dimensionality}), or hybrid embeddings that capture both global statistics and fine-grained dynamics without significantly increasing sequence length.

\paragraph{IoT data enrichment.} 
As previously noted, IoT data alone is insufficient for LLMs to effectively perform real-world tasks. To address this, we enrich the data by incorporating contextual information about the physical world. Specifically, we provide a comprehensive overview of IoT data collection and the integration of physical information. For instance, in human activity recognition (HAR) tasks using inertial measurement unit (IMU) data, including triaxial acceleration and angular velocity from accelerometers and gyroscopes, we detail the data collection process by specifying metadata such as sampling frequency (e.g., 10 Hz), device placement, and measurement units (e.g., gravitational acceleration and radians per second). This method allows LLMs to align the three-axis IMU data with the body's spatial orientations and grasp the physical significance of these values, thereby improving their understanding of the task in the physical world.

\subsubsection*{IoT-oriented Knowledge Retrieval Augmentation}
\label{sec:method:retrieval}
In IoT-sensory task reasoning, the knowledge of LLMs to perform tasks is significant. For example, detecting abnormal heartbeats from electrocardiogram (ECG) data requires interpreting ECG signals and associating them with specific heartbeat states (e.g., premature ventricular contraction), necessitating specialized domain knowledge. Although previous research~\citep{xu2024penetrative} proposes to include specific expert knowledge for specific tasks, the augmentation is task-specific and added manually, which is time-consuming and not scalable. To address this, we enable LLMs with IoT knowledge in an automatic fashion. Inspired by the in-context learning capability of LLMs, we also retrieve task-specific demonstrations, such as question-answer pairs, to guide LLMs in effectively utilizing IoT data for analyzing IoT-sensory tasks.

\paragraph{Construct IoT knowledge base.} 
We firstly construct IoT domain knowledge base and demonstration knowledge base, which will be utilized for retrieving domain knowledge and task-specific demonstrations. For the IoT domain knowledge base, we gather relevant documents (e.g., Wikipedia articles, research papers) through web searches. To ensure comprehensive coverage of IoT data and tasks, we focus on documents encompassing the following themes: (1) IoT data domain knowledge, (2) IoT task domain knowledge, and (3) expert insights on leveraging IoT data for task execution. For instance, in the context of using ECG data to detect abnormal heartbeats, we collect documents that (1) provide detailed explanations of ECG data, such as the significance of various fluctuations, (2) describe the task of heartbeat anomaly detection and the distinctions between normal and abnormal heartbeats, and (3) elucidate the phenomena observed in ECG data when a heartbeat is abnormal. For the demonstration knowledge base, we create task-specific demonstrations authored by human experts or AI models (e.g., ChatGPT). These demonstrations include task description (question), a step-by-step analysis of the task using the provided IoT data (optional), and the final answer to the task.

\paragraph{Embed knowledge base into vector database.} After constructing the knowledge bases, we employ an embedding model (e.g., \href{https://huggingface.co/thenlper/gte-large}{thenlper-gte-large model}) to transform the knowledge base into a vector database. Specifically, we segment documents into smaller chunks (e.g., split by every two sentences). Demonstrations, being shorter in length, are embedded as single chunks. The embedding model then converts these chunks into vector embeddings. Finally, we store these text chunks and corresponding embeddings as key-value pairs, which allows for efficient and scalable search capabilities. To improve the quality of retrieved contents, we also store metadata (e.g. IoT data type for IoT domian knowledge base and task type for demonstration knowledge base) alongside the vector embeddings within the vector database. This approach allows for advanced post-processing techniques, such as metadata filtering~\citep{Poliakov2024MultiMetaRAGIR}, to refine search results and improve task-specific retrieval accuracy.
\paragraph{IoT-oriented retrieval.} In this step, we retrieve relevant knowledge using both IoT data descriptions and task descriptions as query. We employ two types of retrievers for this purpose: (1) sparse keyword-based retrievers and (2) dense embedding-based retrievers. The former identifies similar documents based on shared keywords between the document and the query, excelling in keyword matching but potentially missing semantically similar information. The latter operates at the embedding level, requiring the embedding of IoT data descriptions and task descriptions using the same embedding model as in the previous step to generate query vectors. This kind of retriever is adept at identifying semantically relevant content but may overlook precise interpretations of technical terms. Given that our data descriptions and task descriptions contain numerous numerical data and specialized terminology, we adopt a hybrid search method, which means utilizing both types of retrievers to harness their unique strengths, ensuring the consistent retrieval of highly relevant and context-rich information. Specifically, we retrieve 2n pieces of texts using each type of retriever (n pieces per type) and then apply a re-ranking technique to recalibrate the similarity between the query and documents using ranker models (e.g. \href{https://huggingface.co/BAAI/bge-reranker-base}{bge-reranker-base}), filtering out the top-m most relevant pieces. Ultimately, this process yields pertinent knowledge, encompassing documents with specific domain knowledge and task demonstrations relevant to the task at hand.
It is important to note that we don't need to construct each knowledge base for each task especially, instead, we just need to construct two knowledge bases in total (i.e., one IoT domain knowledge base and one demonstration knowledge base), each of which contains all the domain/demonstration knowledge about the total five tasks. During the retrieval phase, we can easily fetch pertinent knowledge precisely corresponding to the task utilizing metadata (e.g., IoT data type and task type) stored within the bases. For demonstrations, we utilize the one-shot setting, which means we retrieve one example for each category in classification tasks. 

\paragraph{Seamless expansion to additional tasks.} By employing this automatic retrieval method to acquire relevant knowledge, as opposed to the manual design of expert knowledge utilized in Penetrative AI~\citep{xu2024penetrative}, our approach allows for seamless expansion to address additional tasks. This is achieved simply by adding additional documents pertinent to the additional task into the IoT knowledge base, while the rest of the framework's pipeline remains unchanged. In the real world, it's not possible to tailor the expert knowledge in the prompt for thousands of IoT tasks, highlighting our advantages.

\subsubsection*{Prompt Configuration}
In addition to augmenting LLMs' knowledge by providing external documents in the context utilizing the in-context learning capability of LLMs, we further invoke LLMs' internal knowledge by carefully configuring the prompt. Recent research highlights LLMs' robust role-playing abilities~\citep{park2023generative}. We exploit this by assigning specific roles to LLMs for particular tasks. For example, we have LLMs adopt the role of a professional doctor for heartbeat anomaly detection, thereby utilizing their internal domain knowledge. Furthermore, since LLMs' reasoning capabilities can be significantly improved by decomposing problems into smaller parts~\citep{DBLP:conf/nips/Wei0SBIXCLZ22}, we divide the reasoning process into two steps: first prompting LLMs to analyze IoT data and tasks, and then providing the final answer based on this analysis. This approach also allows us to assess the LLMs' understanding of IoT data and their ability to perform IoT tasks through the generated analysis. Finally, we use a prompt template (as illustrated in Fig.~\ref{fig: figure 5 final prompt template} in the appendix) to structure the content discussed. The final prompt is crafted based on the template and then input into a downstream LLM (refer to Fig.~\ref{fig: figure 6 prompt example} in the appendix), which produces the final output, including both analysis and the answer to the specified task.

\subsection*{IoT Dataset Usage}
In our study, we select publicly available IoT datasets for five tasks to ensure fairness. Specifically, for Human Activity Recognition, we use the Smartphone-Based Recognition of Human Activities and Postural Transitions Dataset~\citep{misc_smartphone-based_recognition_of_human_activities_and_postural_transitions_341}. For Industrial Anomaly Detection, we use the Condition Monitoring of Hydraulic Systems Dataset~\citep{misc_condition_monitoring_of_hydraulic_systems_447}. For Heartbeat Anomaly Detection, we employ the MIT-BIH Arrhythmia Database~\citep{932724}. For the Human Sensing task, we utilize a dataset collected with a TP-Link TL-WDR4300 WiFi router operating at 5 GHz with a 40 MHz bandwidth~\citep{zhuravchak2022human}. For the Indoor Localization task, we use a dataset collected in a laboratory environment using an IoT system developed in Varifi~\citep{huang2022varifi}. Due to the substantial scale and class complexity of certain datasets, we use representative subsets—only where necessary—to reduce computational cost, particularly given the high expense of large-scale LLM API calls. This subsampling approach draws on established practices in similar studies (e.g., HarGPT~\citep{ji2024hargpt}) and is carefully designed to preserve the key statistical characteristics of the original datasets, while enabling us to execute nearly 6,000 task instances per model (about 36,000 in total), thereby ensuring both fairness and robustness in subsequent comparisons with baseline methods.

\paragraph{Human Activity Recognition.} We employ the Smartphone-Based Recognition of Human Activities and Postural Transitions Dataset~\citep{misc_smartphone-based_recognition_of_human_activities_and_postural_transitions_341}. This dataset comprises raw IMU data, specifically 3-axial linear acceleration, and 3-axial angular velocity, captured at a sampling rate of 50Hz by the smartphone's accelerometer and gyroscope. The data encompasses twelve distinct activities. To reduce both the sequence length and data complexity, we down-sample the data to 10Hz. Given the challenges associated with multi-class classification for LLM, instead of utilizing all twelve activity categories, we conduct a binary classification task involving the WALKING and STANDING labels, and a ternary classification task with the LYING, WALKING UPSTAIRS, and LIE TO SIT labels. We evaluate the average correction performance using 100 samples per label.

\paragraph{Industrial anomaly detection.} We employ the Condition Monitoring of Hydraulic Systems Dataset~\citep{misc_condition_monitoring_of_hydraulic_systems_447}, which facilitates the assessment of a hydraulic test rig's condition using multi-sensor data, including temperature, cooling power, and efficiency factor series, all experimentally derived from the rig. The dataset categorizes cooler conditions into three severity grades: (1) close to failure; (2) reduced efficiency; and (3) full efficiency. For simplicity, we focus on a binary classification task using only \textquotedblleft close to failure\textquotedblright and \textquotedblleft full efficiency\textquotedblright categories. We evaluate the average correction performance using 700 samples per label.

\paragraph{Heartbeat anomaly detection.} We employ the MIT-BIH Arrhythmia Database~\citep{932724}. This dataset comprises ECG recordings from 48 subjects, each sampled at 360Hz, and categorizes heartbeats into several types, including Normal beat (N), Atrial premature beat (A), and Premature ventricular contraction (V), among others. To reduce the difficulty of the task, we down-sample the signals to 72Hz and focus on a binary classification task using only the Normal beat (N) and Premature ventricular contraction (V) categories. We evaluate the average correction performance using 500 samples per category.

\paragraph{Human sensing task.} We utilize a dataset collected using a TP-Link TL-WDR4300 WiFi router operating at 5 GHz with a 40 MHz bandwidth~\citep{zhuravchak2022human}. The dataset specifically captures the absence of human presence across three different rooms. Each room's environment is carefully monitored to record Channel State Information (CSI) that reflects the presence or absence of occupants, providing a robust basis for occupancy detection tasks. For the experiment, we evaluate the average correction performance using 500 samples per category.

\paragraph{Indoor localization task.} We utilize a dataset collected in a laboratory environment using an IoT system developed in Varifi~\citep{huang2022varifi}. The dataset consists of RSSI signals, the basis for determining human positions within the space. By collecting RSS fingerprints at various reference points, a signal radio map is constructed using a modified Gaussian Process Regression (GPR) method. This approach allows us to estimate the RSS distribution at any given location, providing a reliable means of localizing human presence in the environment. A total of 2,000 samples were utilized to compute the evaluation metrics mentioned in the experiment.

\subsection*{Experimental Cost}
The most closely related works to ours are Penetrative AI and HarGPT, both of which rely on hand-crafted IoT domain knowledge designed by experts. This manual process requires precise, task-specific expertise and does not scale well to additional domains. As a result, adapting to different IoT tasks demands substantial effort, with each task necessitating careful modification of its corresponding knowledge base. Consequently, these approaches typically address only one or a few tasks.
In contrast, our method eliminates the need for manual, task-specific knowledge engineering. Instead, we construct a knowledge base by collecting relevant domain-specific articles from online resources and literature. The RAG module in our framework then automatically retrieves the required knowledge for each task. Extending to an additional IoT task merely involves adding relevant documents to the repository, without incurring additional expert labor. Compared to Penetrative AI and HarGPT, our approach completely removes human expert costs, with the primary experimental expense limited to API usage fees for closed-source models.

For open-source LLMs such as LLama2-7B, inference was conducted locally, utilizing an A6000 GPU equipped with 48GB of VRAM. In contrast, for closed-source models like GPT-4o-mini, gemini-pro, and claude, inference was performed via their respective official APIs, where costs are determined by the number of tokens processed. In our experiments, the average inference cost per 100 samples was \$0.26 for GPT-3.5-turbo, \$0.10 for GPT-4o-mini, \$1.00 for gemini-pro, and \$2.50 for claude-3.5-sonnet. Based on the total number of samples processed, the overall API expenditure amounted to approximately \$450.

Regarding inference time, we did not optimize for speed and inference time scales with input length. On average, finishing a single task instance takes 7.45 seconds in our framework. This latency is acceptable for the five real-world IoT tasks in our benchmark, as real-time feedback is not required. By contrast, the Penetrative AI method completes a task in approximately 5.51 seconds. The longer runtime of our framework is primarily due to the retrieval of deeper expert knowledge, which increases the input length. Experimental results indicate that our method substantially outperforms Penetrative AI in performance. Consequently, there is an inherent trade-off between performance and inference time. For deployment on resource-constrained IoT edge devices or tasks requiring rapid feedback, the core LLM can be replaced with lighter alternatives to reduce latency and resource consumption. For example, in our experiments the core model (e.g., Mistral-7B) can be substituted with a lightweight model, including models produced via knowledge distillation or through pruning and quantization ~\citep{behdin2025efficient, fang2025knowledge, douzandeh2025comparative, lang2024comprehensive}. Nevertheless, efficient inference and task performance are in tension, requiring a practical balance determined by the application's requirements.

\newpage


\section*{RESOURCE AVAILABILITY}


\subsection*{Lead contact}


Requests for further information and resources should be directed to and will be fulfilled by the lead contact, Jianfei Yang. (yang0478@ntu.edu.sg).

\subsection*{Materials availability}


This study did not generate new materials.

\subsection*{Data and code availability}


\begin{itemize}
    \item The datasets used in this paper are publicly available from the following sources. \\Smartphone-Based Recognition of Human Activities and Postural Transitions Dataset is available from \href{https://doi.org/10.24432/C54G7M}{this link}. Condition Monitoring of Hydraulic Systems Dataset is available from \href{https://doi.org/10.24432/C5CW21}{this link}. MIT-BIH Arrhythmia Database is available on \href{https://physionet.org/content/mitdb/1.0.0/}{PhysioNet} . Dataset for human sensing task is available at \href{https://doi.org/10.6084/m9Ω.figshare.14386892.v1}{this link}. The dataset for the WiFi-based indoor localization task was collected in a laboratory environment using an IoT system developed in Varifi \citep{huang2022varifi}. For access to the dataset, please contact the lead author.
    \item The codes generated during this study have been deposited in \href{https://github.com/Morpheus-An/IoT-Agent.git}{Github} and \href{https://doi.org/10.5281/zenodo.17309286}{Zenodo}.\citep{code_for_IoT-agent} 

    \item Any additional information required to reanalyze the data reported in this paper is available from the lead contact upon request.    
\end{itemize}

\section*{ACKNOWLEDGMENTS}


This work is supported by a Start-up Grant from Nanyang Technological University and jointly funded by the Singapore Ministry of Education (MOE) under a Tier-1 research grant.

\section*{AUTHOR CONTRIBUTIONS}


Conceptualization, T.A. and J.Y.; methodology, T.A. and J.Y.; investigation, T.A. and Y.Z.; validation, T.A.; data curation, T.A. and Y.Z.; writing - original draft, T.A.; writing - review \& editing, H.Z. and J.Y.; funding acquisition, J.Y.

\section*{DECLARATION OF INTERESTS}

The authors declare no competing interests.

\newpage

\clearpage


\bibliography{references}

@misc{code_for_IoT-agent,
  author       = {An,T.},
  title        = {Morpheus-An/IoT-Agent: cell patterns (v1.0.0)},
  month        = oct,
  year         = 2025,
  publisher    = {Zenodo},
  version      = {v1.0.0},
  note         = {\href{https://doi.org/10.5281/zenodo.17310907}{https://doi.org/10.5281/zenodo.17310907}}
}

@article{zhou2024larger,
  title={Larger and more instructable language models become less reliable},
  author={Zhou, Lexin and Schellaert, Wout and Mart{\'\i}nez-Plumed, Fernando and Moros-Daval, Yael and Ferri, C{\`e}sar and Hern{\'a}ndez-Orallo, Jos{\'e}},
  journal={Nature},
  pages={1--8},
  year={2024},
  publisher={Nature Publishing Group UK London}
}

@article{fedorenko2024language,
  title={Language is primarily a tool for communication rather than thought},
  author={Fedorenko, Evelina and Piantadosi, Steven T and Gibson, Edward AF},
  journal={Nature},
  volume={630},
  number={8017},
  pages={575--586},
  year={2024},
  publisher={Nature Publishing Group UK London}
}

@article{Zhang2024LargeLM,
  title={Large Language Models for Time Series: A Survey},
  author={Xiyuan Zhang and Ranak Roy Chowdhury and Rajesh K. Gupta and Jingbo Shang},
  journal={Preprint at arXiv},
  year={2024},
  note={\href{https://doi.org/10.48550/arXiv.2402.01801}{https://doi.org/10.48550/arXiv.2402.01801}}
}

@article{DBLP:journals/jamia/SpathisK24,
  author       = {Dimitris Spathis and
                  Fahim Kawsar},
  title        = {\emph{The first step is the hardest}: pitfalls of representing and
                  tokenizing temporal data for large language models},
  journal      = {J. Am. Medical Informatics Assoc.},
  volume       = {31},
  number       = {9},
  pages        = {2151--2158},
  year         = {2024}
}

@article{madakam2015internet,
  title={Internet of Things (IoT): A literature review},
  author={Madakam, Somayya and Ramaswamy, Ramya and Tripathi, Siddharth},
  journal={Journal of Computer and Communications},
  volume={3},
  number={5},
  pages={164--173},
  year={2015},
  publisher={Scientific Research Publishing}
}

@article{zhang2024visionlanguage,
  author    = {Zhang, Jingyi and Huang, Jiaxing and Jin, Sheng and Lu, Shijian},
  title     = {Vision-Language Models for Vision Tasks: A Survey},
  journal   = {IEEE Transactions on Pattern Analysis and Machine Intelligence},
  volume    = {46},
  number    = {8},
  year      = {2024},
  note       = {\href{https://doi.org/10.1109/TPAMI.2024.3369699}{https://doi.org/10.1109/TPAMI.2024.3369699}}
}

@article{gruver2024large,
  title={Large language models are zero-shot time series forecasters},
  author={Gruver, Nate and Finzi, Marc and Qiu, Shikai and Wilson, Andrew G},
  journal={Advances in Neural Information Processing Systems},
  volume={36},
  year={2024}
}

@inproceedings{xu-etal-2024-penetrative,
    title = "Penetrative {AI}: Making {LLM}s Comprehend the Physical World",
    author = "Xu, Huatao  and
      Han, Liying  and
      Yang, Qirui  and
      Li, Mo  and
      Srivastava, Mani",
    editor = "Ku, Lun-Wei  and
      Martins, Andre  and
      Srikumar, Vivek",
    booktitle = "Findings of the Association for Computational Linguistics ACL 2024",
    month = aug,
    year = "2024",
    address = "Bangkok, Thailand and virtual meeting",
    publisher = "Association for Computational Linguistics",
    note = {\href{https://aclanthology.org/2024.findings-acl.437}{https://aclanthology.org/2024.findings-acl.437}},
    pages = "7324--7341",
}

@techreport{radford2018improving,
  title        = {Improving Language Understanding by Generative Pre-Training},
  author       = {Radford, Alec and Narasimhan, Karthik and Salimans, Tim and Sutskever, Ilya},
  year         = {2018},
  institution  = {OpenAI},
note         = {\href{https://cdn.openai.com/research-covers/language-unsupervised/language_understanding_paper.pdf}{https://cdn.openai.com/research-covers/language-unsupervised/language\_understanding\_paper.pdf}}
}

@article{cui2023llmind,
  title={Llmind: Orchestrating ai and iot with llms for complex task execution},
  author={Cui, Hongwei and Du, Yuyang and Yang, Qun and Shao, Yulin and Liew, Soung Chang},
  journal={Preprint at arXiv},
  year={2023},
note={\href{https://doi.org/10.48550/arXiv.2312.09007}{https://doi.org/10.48550/arXiv.2312.09007}}
}

@inproceedings{du2023space,
  title={Space Brain: An AI Autonomous Spatial Decision System},
  author={Du, Jiachen and Jia, Boyang and Fu, Xinyi},
  booktitle={CAAI International Conference on Artificial Intelligence},
  pages={61--67},
  year={2023},
  organization={Springer}
}

@inproceedings{li2023chatiot,
  title={ChatIoT: Zero-code Generation of Trigger-action Based IoT Programs with ChatGPT},
  author={Li, Fu and Huang, Jiaming and Gao, Yi and Dong, Wei},
  booktitle={Proceedings of the 7th Asia-Pacific Workshop on Networking},
  pages={219--220},
  year={2023}
}

@article{churchland1988perspectives,
  title={Perspectives on cognitive neuroscience},
  author={Churchland, Patricia S and Sejnowski, Terrence J},
  journal={Science},
  volume={242},
  number={4879},
  pages={741--745},
  year={1988},
  publisher={American Association for the Advancement of Science}
}

@article{saxe2009brain,
  title={Brain regions for perceiving and reasoning about other people in school-aged children},
  author={Saxe, Rebecca R and Whitfield-Gabrieli, Susan and Scholz, Jonathan and Pelphrey, Kevin A},
  journal={Child development},
  volume={80},
  number={4},
  pages={1197--1209},
  year={2009},
  publisher={Wiley Online Library}
}

@article{grezes2001does,
  title={Does perception of biological motion rely on specific brain regions?},
  author={Grezes, Julie and Fonlupt, Pierre and Bertenthal, Bennett and Delon-Martin, Chantal and Segebarth, Christoph and Decety, Jean},
  journal={Neuroimage},
  volume={13},
  number={5},
  pages={775--785},
  year={2001},
  publisher={Elsevier}
}

@article{hobeika2016general,
  title={General and specialized brain correlates for analogical reasoning: A meta-analysis of functional imaging studies},
  author={Hobeika, Lucie and Diard-Detoeuf, Capucine and Garcin, B{\'e}atrice and Levy, Richard and Volle, Emmanuelle},
  journal={Human brain mapping},
  volume={37},
  number={5},
  pages={1953--1969},
  year={2016},
  publisher={Wiley Online Library}
}

@inproceedings{dosovitskiy2021image,
  author       = {Alexey Dosovitskiy and Lucas Beyer and Alexander Kolesnikov and Dirk Weissenborn 
                  and Xiaohua Zhai and Thomas Unterthiner and Mostafa Dehghani and Matthias Minderer 
                  and Georg Heigold and Sylvain Gelly and Jakob Uszkoreit and Neil Houlsby},
  title        = {An Image Is Worth 16×16 Words: Transformers for Image Recognition at Scale},
  booktitle    = {Proceedings of the International Conference on Learning Representations (ICLR) 2021},
  year         = {2021},
  note         = {\href{https://openreview.net/forum?id=YicbFdNTTy}{https://openreview.net/forum?id=YicbFdNTTy}},
}

@inproceedings{DBLP:conf/nips/Wei0SBIXCLZ22,
  author       = {Jason Wei and
                  Xuezhi Wang and
                  Dale Schuurmans and
                  Maarten Bosma and
                  Brian Ichter and
                  Fei Xia and
                  Ed H. Chi and
                  Quoc V. Le and
                  Denny Zhou},
  title        = {Chain-of-Thought Prompting Elicits Reasoning in Large Language Models},
  booktitle    = {NeurIPS},
  year         = {2022}
}

@article{liu2024world,
  title={World Model on Million-Length Video And Language With RingAttention},
  author={Liu, Hao and Yan, Wilson and Zaharia, Matei and Abbeel, Pieter},
  journal={Preprint at arXiv},
  year={2024},
note={\href{https://doi.org/10.48550/arXiv.2402.08268}{https://doi.org/10.48550/arXiv.2402.08268}}
}

@article{garrido2024learning,
  title={Learning and Leveraging World Models in Visual Representation Learning},
  author={Garrido, Quentin and Assran, Mahmoud and Ballas, Nicolas and Bardes, Adrien and Najman, Laurent and LeCun, Yann},
  journal={Preprint at arXiv},
  year={2024},
note={\href{https://doi.org/10.48550/arXiv.2403.00504}{https://doi.org/10.48550/arXiv.2403.00504}}
}

@article{dawid2023introduction,
  title={Introduction to latent variable energy-based models: A path towards autonomous machine intelligence},
  author={Dawid, Anna and LeCun, Yann},
  journal={Preprint at arXiv},
  year={2023},
note={\href{https://doi.org/10.48550/arXiv.2306.02572}{https://doi.org/10.48550/arXiv.2306.02572}}
}

@article{zhang2023meta,
  title={Meta-transformer: A unified framework for multimodal learning},
  author={Zhang, Yiyuan and Gong, Kaixiong and Zhang, Kaipeng and Li, Hongsheng and Qiao, Yu and Ouyang, Wanli and Yue, Xiangyu},
  journal={Preprint at arXiv},
  year={2023},
note={\href{https://doi.org/10.48550/arXiv.2307.10802}{https://doi.org/10.48550/arXiv.2307.10802}}
}

@article{videoworldsimulators2024,
  title={Video generation models as world simulators},
  author={Tim Brooks and Bill Peebles and Connor Holmes and Will DePue and Yufei Guo and Li Jing and David Schnurr and Joe Taylor and Troy Luhman and Eric Luhman and Clarence Ng and Ricky Wang and Aditya Ramesh},
  year={2024},
  note={\href{https://openai.com/research/video-generation-models-as-world-simulators}{https://openai.com/research/video-generation-models-as-world-simulators}},
}

@misc{misc_condition_monitoring_of_hydraulic_systems_447,
  author       = {Helwig,Nikolai and Pignanelli,Eliseo and Schtze,Andreas},
  title        = {Condition monitoring of hydraulic systems},
  year         = {2018},
  howpublished = {UCI Machine Learning Repository},
  note         = {\href{https://doi.org/10.24432/C5CW21}{https://doi.org/10.24432/C5CW21}}
}

@ARTICLE{932724,
  author={Moody, G.B. and Mark, R.G.},
  journal={IEEE Engineering in Medicine and Biology Magazine}, 
  title={The impact of the MIT-BIH Arrhythmia Database}, 
  year={2001},
  volume={20},
  number={3},
  pages={45-50},
  doi={10.1109/51.932724}
}

@misc{misc_smartphone-based_recognition_of_human_activities_and_postural_transitions_341,
  author={Reyes-Ortiz,Jorge and Anguita,Davide and Oneto,Luca and Parra,Xavier},
  title={Smartphone-Based Recognition of Human Activities and Postural Transitions},
  year={2015},
  howpublished={UCI Machine Learning Repository},
  note={\href{https://doi.org/10.24432/C54G7M}{https://doi.org/10.24432/C54G7M}}
}

@article{ji2024hargpt,
  title={HARGPT: Are LLMs Zero-Shot Human Activity Recognizers?},
  author={Ji, Sijie and Zheng, Xinzhe and Wu, Chenshu},
  journal={Preprint at arXiv},
  year={2024},
note={\href{https://doi.org/10.48550/arXiv.2403.02727}{https://doi.org/10.48550/arXiv.2403.02727}}
}

@article{Poliakov2024MultiMetaRAGIR,
  title={Multi-Meta-RAG: Improving RAG for Multi-Hop Queries using Database Filtering with LLM-Extracted Metadata},
  author={Mykhailo Poliakov and Nadiya Shvai},
  journal={Preprint at arXiv},
  year={2024},
  note={\href{https://doi.org/10.48550/arXiv.2406.13213}{https://doi.org/10.48550/arXiv.2406.13213}}
}

@inproceedings{park2023generative,
  title={Generative agents: Interactive simulacra of human behavior},
  author={Park, Joon Sung and O'Brien, Joseph and Cai, Carrie Jun and Morris, Meredith Ringel and Liang, Percy and Bernstein, Michael S},
  booktitle={Proceedings of the 36th annual acm symposium on user interface software and technology},
  pages={1--22},
  year={2023}
}

@inproceedings{xu2024penetrative,
  title={Penetrative ai: Making llms comprehend the physical world},
  author={Xu, Huatao and Han, Liying and Yang, Qirui and Li, Mo and Srivastava, Mani},
  booktitle={Proceedings of the 25th International Workshop on Mobile Computing Systems and Applications},
  pages={1--7},
  year={2024}
}

@article{yang2024mm,
  title={Mm-fi: Multi-modal non-intrusive 4d human dataset for versatile wireless sensing},
  author={Yang, Jianfei and Huang, He and Zhou, Yunjiao and Chen, Xinyan and Xu, Yuecong and Yuan, Shenghai and Zou, Han and Lu, Chris Xiaoxuan and Xie, Lihua},
  journal={Advances in Neural Information Processing Systems},
  volume={36},
  year={2024}
}

@article{blattmann2023stable,
  title={Stable video diffusion: Scaling latent video diffusion models to large datasets},
  author={Blattmann, Andreas and Dockhorn, Tim and Kulal, Sumith and Mendelevitch, Daniel and Kilian, Maciej and Lorenz, Dominik and Levi, Yam and English, Zion and Voleti, Vikram and Letts, Adam and others},
  journal={Preprint at arXiv},
  year={2023},
note = {\href{https://doi.org/10.48550/arXiv.2311.15127}{https://doi.org/10.48550/arXiv.2311.15127}}
}

@inproceedings{peebles2023scalable,
  title={Scalable diffusion models with transformers},
  author={Peebles, William and Xie, Saining},
  booktitle={Proceedings of the IEEE/CVF International Conference on Computer Vision},
  pages={4195--4205},
  year={2023}
}

@article{huang2023survey,
  title={A survey on hallucination in large language models: Principles, taxonomy, challenges, and open questions},
  author={Huang, Lei and Yu, Weijiang and Ma, Weitao and Zhong, Weihong and Feng, Zhangyin and Wang, Haotian and Chen, Qianglong and Peng, Weihua and Feng, Xiaocheng and Qin, Bing and others},
  journal={Preprint at arXiv},
  year={2023},
note={\href{https://doi.org/10.1145/3703155}{https://doi.org/10.1145/3703155}}
}

@article{alkaissi2023artificial,
  title={Artificial hallucinations in ChatGPT: implications in scientific writing},
  author={Alkaissi, Hussam and McFarlane, Samy I},
  journal={Cureus},
  volume={15},
  number={2},
  year={2023},
  publisher={Cureus}
}

@article{achiam2023gpt,
  title={Gpt-4 technical report},
  author={Achiam, Josh and Adler, Steven and Agarwal, Sandhini and Ahmad, Lama and Akkaya, Ilge and Aleman, Florencia Leoni and Almeida, Diogo and Altenschmidt, Janko and Altman, Sam and Anadkat, Shyamal and others},
  journal={Preprint at arXiv},
  year={2023},
note = {\href{https://doi.org/10.48550/arXiv.2303.08774}{https://doi.org/10.48550/arXiv.2303.08774}}
}

@article{Ho2020DenoisingDP,
  title={Denoising Diffusion Probabilistic Models},
  author={Jonathan Ho and Ajay Jain and P. Abbeel},
  journal={Preprint at arXiv},
  year={2020},
  note={\href{https://doi.org/10.48550/arXiv.2006.11239}{https://doi.org/10.48550/arXiv.2006.11239}}
}

@article{brown2020language,
  title={Language models are few-shot learners},
  author={Brown, Tom and Mann, Benjamin and Ryder, Nick and Subbiah, Melanie and Kaplan, Jared D and Dhariwal, Prafulla and Neelakantan, Arvind and Shyam, Pranav and Sastry, Girish and Askell, Amanda and others},
  journal={Advances in neural information processing systems},
  volume={33},
  pages={1877--1901},
  year={2020}
}

@article{radford2019language,
  title={Language models are unsupervised multitask learners},
  author={Radford, Alec and Wu, Jeffrey and Child, Rewon and Luan, David and Amodei, Dario and Sutskever, Ilya and others},
  journal={OpenAI blog},
  volume={1},
  number={8},
  pages={9},
  year={2019}
}

@article{zhuravchak2022human,
  title={Human activity recognition based on wi-fi csi data-a deep neural network approach},
  author={Zhuravchak, Andrii and Kapshii, Oleg and Pournaras, Evangelos},
  journal={Procedia Computer Science},
  volume={198},
  pages={59--66},
  year={2022},
  publisher={Elsevier}
}

@article{huang2022varifi,
  title={VariFi: Variational Inference for Indoor Pedestrian Localization and Tracking Using IMU and WiFi RSS},
  author={Huang, He and Yang, Jianfei and Fang, Xu and Jiang, Hao and Xie, Lihua},
  journal={IEEE Internet of Things Journal},
  volume={10},
  number={10},
  pages={9049--9061},
  year={2022},
  publisher={IEEE}
}

@article{kim2024openvla,
  title={Openvla: An open-source vision-language-action model},
  author={Kim, Moo Jin and Pertsch, Karl and Karamcheti, Siddharth and Xiao, Ted and Balakrishna, Ashwin and Nair, Suraj and Rafailov, Rafael and Foster, Ethan and Lam, Grace and Sanketi, Pannag and others},
  journal={Preprint at arXiv},
  year={2024},
note={\href{https://doi.org/10.48550/arXiv.2406.09246}{https://doi.org/10.48550/arXiv.2406.09246}}
}

@article{mu2023embodiedgpt,
  title={Embodiedgpt: Vision-language pre-training via embodied chain of thought},
  author={Mu, Yao and Zhang, Qinglong and Hu, Mengkang and Wang, Wenhai and Ding, Mingyu and Jin, Jun and Wang, Bin and Dai, Jifeng and Qiao, Yu and Luo, Ping},
  journal={Advances in Neural Information Processing Systems},
  volume={36},
  pages={25081--25094},
  year={2023}
}

@article{wang2024large,
  title={Large Language Models for Robotics: Opportunities, Challenges, and Perspectives},
  author={Wang, Jiaqi and Wu, Zihao and Li, Yiwei and Jiang, Hanqi and Shu, Peng and Shi, Enze and Hu, Huawen and Ma, Chong and Liu, Yiheng and Wang, Xuhui and others},
  journal={Preprint at arXiv},
  year={2024},
note = {\href{https://doi.org/10.48550/arXiv.2401.04334}{https://doi.org/10.48550/arXiv.2401.04334}}
}

@article{yang2025fuzzcode,
  title={FuzzCode: Code Large Language Model-Based Fuzz Testing for Industrial IoT Programs},
  author={Yang, Liqun and Wei, Chaoren and Yang, Jian and Xia, Wanxu and Yang, Yuze and Luo, Yang and Niyato, Dusit and Sun, Liang and Liu, Zhiquan},
  journal={IEEE Internet of Things Journal},
  year={2025},
  publisher={IEEE}
}

@article{xia2024fcllm,
  title={Fcllm-dt: Enpowering federated continual learning with large language models for digital twin-based industrial iot},
  author={Xia, Yingjie and Chen, Yuhan and Zhao, Yunxiao and Kuang, Li and Liu, Xuejiao and Hu, Ji and Liu, Zhiquan},
  journal={IEEE Internet of Things Journal},
  year={2024},
  publisher={IEEE}
}

@article{behdin2025efficient,
  title={Efficient AI in Practice: Training and Deployment of Efficient LLMs for Industry Applications},
  author={Behdin, Kayhan and Dai, Yun and Fatahibaarzi, Ata and Gupta, Aman and Song, Qingquan and Tang, Shao and Sang, Hejian and Dexter, Gregory and Zhu, Sirou and Zhu, Siyu and others},
  journal={Preprint at arXiv},
  year={2025},
note={\href{https://doi.org/10.48550/arXiv.2502.14305}{https://doi.org/10.48550/arXiv.2502.14305}}
}

@article{fang2025knowledge,
  title={Knowledge distillation and dataset distillation of large language models: Emerging trends, challenges, and future directions},
  author={Fang, Luyang and Yu, Xiaowei and Cai, Jiazhang and Chen, Yongkai and Wu, Shushan and Liu, Zhengliang and Yang, Zhenyuan and Lu, Haoran and Gong, Xilin and Liu, Yufang and others},
  journal={Preprint at arXiv},
  year={2025},
note={\href{https://doi.org/10.48550/arXiv.2504.14772}{https://doi.org/10.48550/arXiv.2504.14772}}
}

@article{douzandeh2025comparative,
  title={A comparative study of neural network pruning strategies for industrial applications},
  author={Douzandeh Zenoozi, Amirhossein and Erhan, Laura and Liotta, Antonio and Cavallaro, Lucia},
  journal={Frontiers in Computer Science},
  volume={7},
  pages={1563942},
  year={2025},
  publisher={Frontiers Media SA}
}

@inproceedings{lang2024comprehensive,
  title={A comprehensive study on quantization techniques for large language models},
  author={Lang, Jiedong and Guo, Zhehao and Huang, Shuyu},
  booktitle={2024 4th International Conference on Artificial Intelligence, Robotics, and Communication (ICAIRC)},
  pages={224--231},
  year={2024},
  organization={IEEE}
}

@inproceedings{howard2019searching,
  title={Searching for mobilenetv3},
  author={Howard, Andrew and Sandler, Mark and Chu, Grace and Chen, Liang-Chieh and Chen, Bo and Tan, Mingxing and Wang, Weijun and Zhu, Yukun and Pang, Ruoming and Vasudevan, Vijay and others},
  booktitle={Proceedings of the IEEE/CVF international conference on computer vision},
  pages={1314--1324},
  year={2019}
}

@inproceedings{tan2019efficientnet,
  title={Efficientnet: Rethinking model scaling for convolutional neural networks},
  author={Tan, Mingxing and Le, Quoc},
  booktitle={International conference on machine learning},
  pages={6105--6114},
  year={2019},
  organization={PMLR}
}

@article{qi2017pointnet++,
  title={Pointnet++: Deep hierarchical feature learning on point sets in a metric space},
  author={Qi, Charles Ruizhongtai and Yi, Li and Su, Hao and Guibas, Leonidas J},
  journal={Advances in neural information processing systems},
  volume={30},
  year={2017}
}

@inproceedings{hu2020randla,
  title={Randla-net: Efficient semantic segmentation of large-scale point clouds},
  author={Hu, Qingyong and Yang, Bo and Xie, Linhai and Rosa, Stefano and Guo, Yulan and Wang, Zhihua and Trigoni, Niki and Markham, Andrew},
  booktitle={Proceedings of the IEEE/CVF conference on computer vision and pattern recognition},
  pages={11108--11117},
  year={2020}
}

@article{abdi2010principal,
  title={Principal component analysis},
  author={Abdi, Herv{\'e} and Williams, Lynne J},
  journal={Wiley interdisciplinary reviews: computational statistics},
  volume={2},
  number={4},
  pages={433--459},
  year={2010},
  publisher={Wiley Online Library}
}

@article{tharwat2017linear,
  title={Linear discriminant analysis: A detailed tutorial},
  author={Tharwat, Alaa and Gaber, Tarek and Ibrahim, Abdelhameed and Hassanien, Aboul Ella},
  journal={AI communications},
  volume={30},
  number={2},
  pages={169--190},
  year={2017},
  publisher={SAGE Publications Sage UK: London, England}
}

@article{cieslak2020t,
  title={t-Distributed Stochastic Neighbor Embedding (t-SNE): A tool for eco-physiological transcriptomic analysis},
  author={Cieslak, Matthew C and Castelfranco, Ann M and Roncalli, Vittoria and Lenz, Petra H and Hartline, Daniel K},
  journal={Marine genomics},
  volume={51},
  pages={100723},
  year={2020},
  publisher={Elsevier}
}

@article{mcinnes2018umap,
  title={Umap: Uniform manifold approximation and projection for dimension reduction},
  author={McInnes, Leland and Healy, John and Melville, James},
  journal={Preprint at arXiv},
  year={2018},
note={\href{https://doi.org/10.48550/arXiv.1802.03426}{https://doi.org/10.48550/arXiv.1802.03426}}
}

@inproceedings{liu2021swin,
  title={Swin transformer: Hierarchical vision transformer using shifted windows},
  author={Liu, Ze and Lin, Yutong and Cao, Yue and Hu, Han and Wei, Yixuan and Zhang, Zheng and Lin, Stephen and Guo, Baining},
  booktitle={Proceedings of the IEEE/CVF international conference on computer vision},
  pages={10012--10022},
  year={2021}
}

@inproceedings{han2024onellm,
  title={Onellm: One framework to align all modalities with language},
  author={Han, Jiaming and Gong, Kaixiong and Zhang, Yiyuan and Wang, Jiaqi and Zhang, Kaipeng and Lin, Dahua and Qiao, Yu and Gao, Peng and Yue, Xiangyu},
  booktitle={Proceedings of the IEEE/CVF Conference on Computer Vision and Pattern Recognition},
  pages={26584--26595},
  year={2024}
}

@article{krishnamurthi2020overview,
  title={An overview of IoT sensor data processing, fusion, and analysis techniques},
  author={Krishnamurthi, Rajalakshmi and Kumar, Adarsh and Gopinathan, Dhanalekshmi and Nayyar, Anand and Qureshi, Basit},
  journal={Sensors},
  volume={20},
  number={21},
  pages={6076},
  year={2020},
  publisher={MDPI}
}

@article{kim2022anomaly,
  title={Anomaly detection with feature extraction based on machine learning using hydraulic system IoT sensor data},
  author={Kim, Doyun and Heo, Tae-Young},
  journal={Sensors},
  volume={22},
  number={7},
  pages={2479},
  year={2022},
  publisher={MDPI}
}

@article{wu2018feature,
  title={A feature-based learning system for Internet of Things applications},
  author={Wu, Dapeng and Shi, Hang and Wang, Honggang and Wang, Ruyan and Fang, Hua},
  journal={IEEE Internet of things Journal},
  volume={6},
  number={2},
  pages={1928--1937},
  year={2018},
  publisher={IEEE}
}

@article{singh2024tokenization,
  title={Tokenization counts: the impact of tokenization on arithmetic in frontier llms},
  author={Singh, Aaditya K and Strouse, DJ},
  journal={Preprint at arXiv},
  year={2024},
note={\href{https://doi.org/10.48550/arXiv.2402.14903}{https://doi.org/10.48550/arXiv.2402.14903}}
}

@article{lin2007experiencing,
  title={Experiencing SAX: a novel symbolic representation of time series},
  author={Lin, Jessica and Keogh, Eamonn and Wei, Li and Lonardi, Stefano},
  journal={Data Mining and knowledge discovery},
  volume={15},
  number={2},
  pages={107--144},
  year={2007},
  publisher={Springer}
}

@article{keogh2001dimensionality,
  title={Dimensionality reduction for fast similarity search in large time series databases},
  author={Keogh, Eamonn and Chakrabarti, Kaushik and Pazzani, Michael and Mehrotra, Sharad},
  journal={Knowledge and information Systems},
  volume={3},
  number={3},
  pages={263--286},
  year={2001},
  publisher={Springer}
}
\newpage

\section*{APPENDIX}

\begin{figure*}[h!]
\begin{center}
\includegraphics[width=1.0\textwidth]{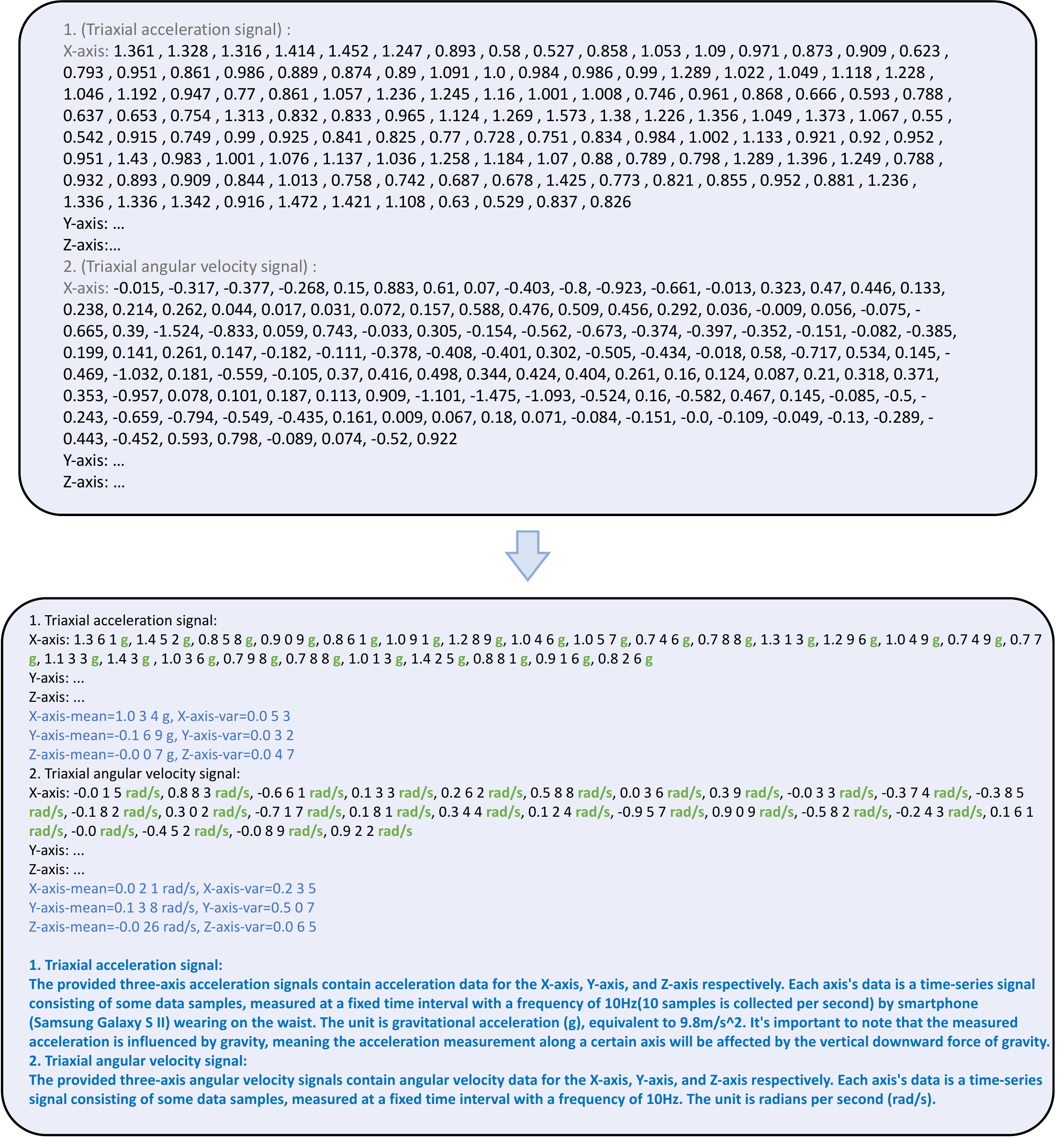}
\end{center}
\caption{During IoT data simplification and enrichment stage, raw IoT data is transformed into IoT data description, which is easier to understand by LLMs. Raw IoT data is enriched with descriptive metadata, including natural language expressions of implicit physical information like units. Specialized tokenization techniques and extraction of temporal or frequency domain features further enhance LLMs' understanding of numerical and time-series data. These improvements make IoT data more accessible and interpretable for LLMs, facilitating its use in real-world applications.}
\label{fig: figure 4 data preprocess}
\end{figure*}

\begin{figure*}[htbp]
\begin{center}
\includegraphics[width=1.0\textwidth]{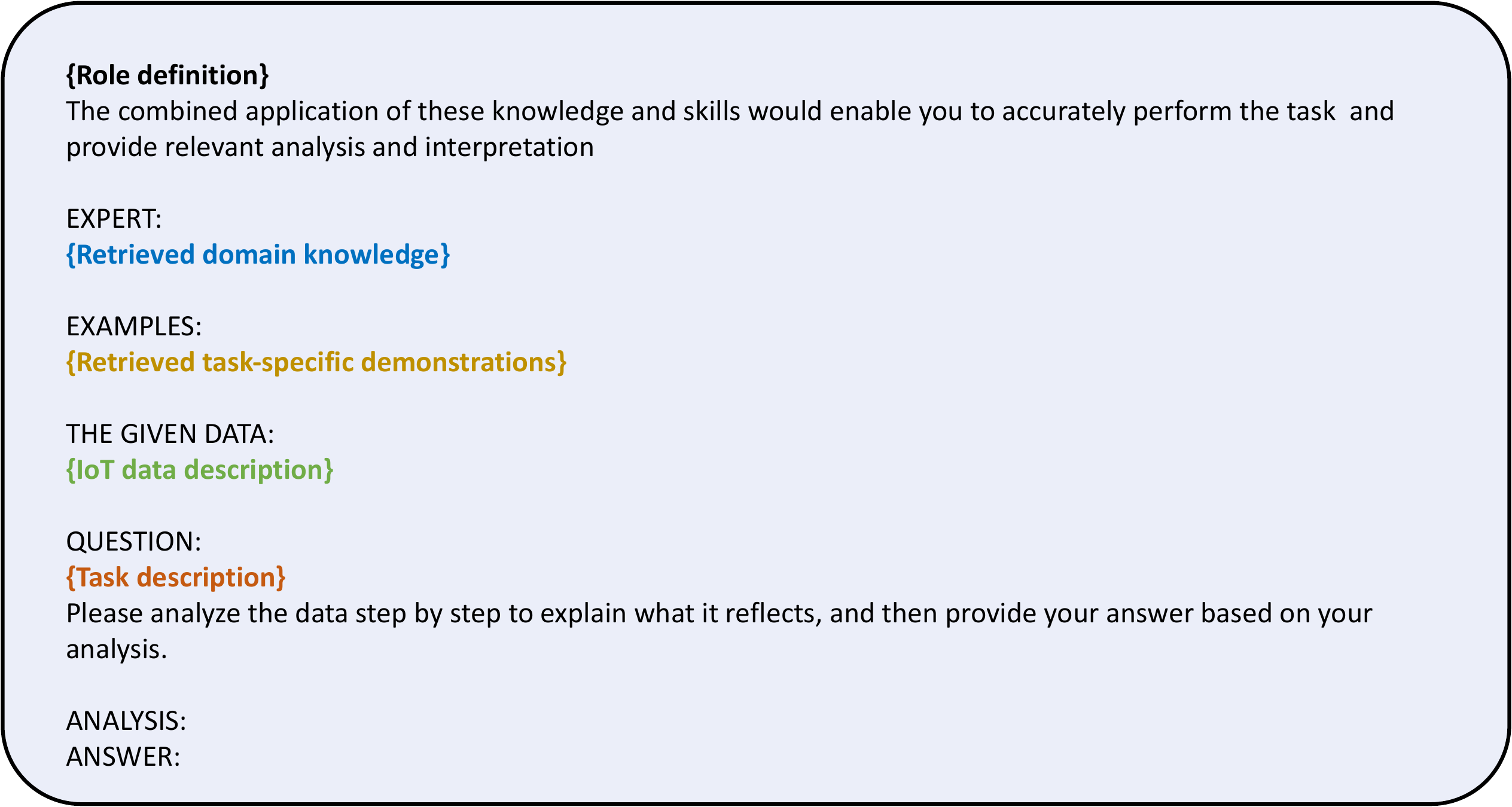}
\end{center}
\caption{Final prompt template.}
\label{fig: figure 5 final prompt template}
\end{figure*}

\begin{figure*}[htbp]
\begin{center}
\includegraphics[width=0.85\textwidth]{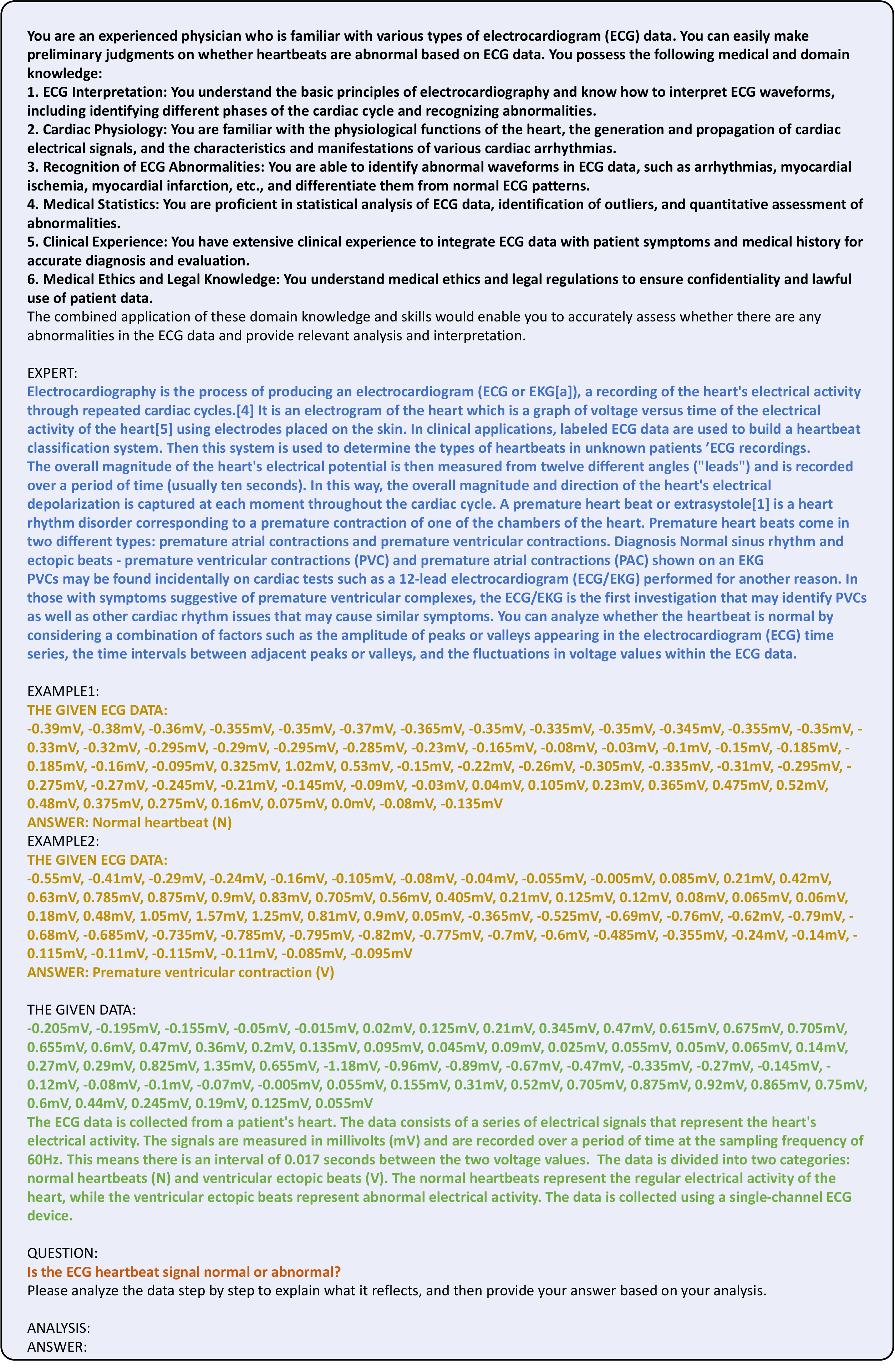}
\end{center}
\caption{Final prompt example for heartbeat anomaly detection. Note that role description is generated automatically by AI models (e.g., ChatGPT).}
\label{fig: figure 6 prompt example}
\end{figure*}

\begin{figure*}[!h]
\begin{center}
\includegraphics[width=1\textwidth]{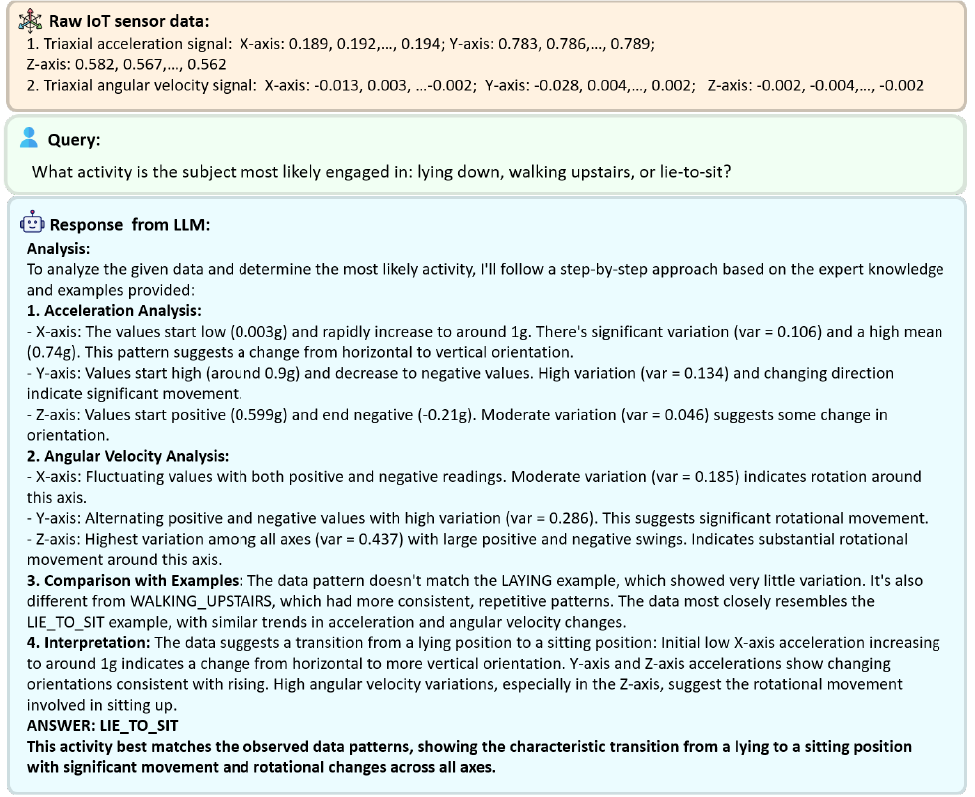}
\end{center}
\caption{Response example of LLM for human activity recognition.}
\label{fig: HAR}
\end{figure*}

\begin{figure*}[!b]
\begin{center}
\includegraphics[width=1\textwidth]{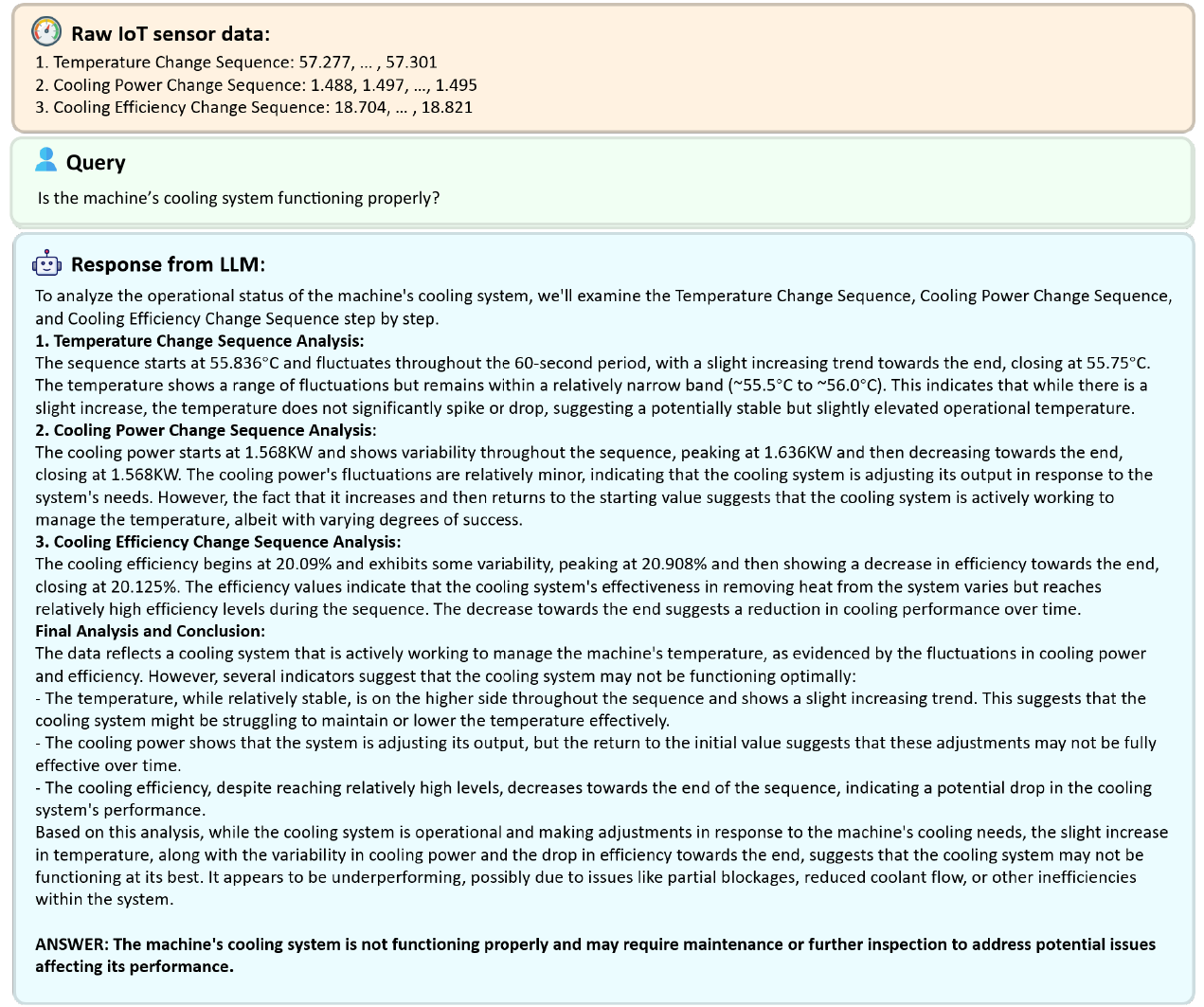}
\end{center}
\caption{Response example of LLM for industrial anomaly detection.}
\label{fig: anomaly detection}
\end{figure*}

\clearpage
\begin{figure*}[b]
\begin{center}
\includegraphics[width=1\textwidth]{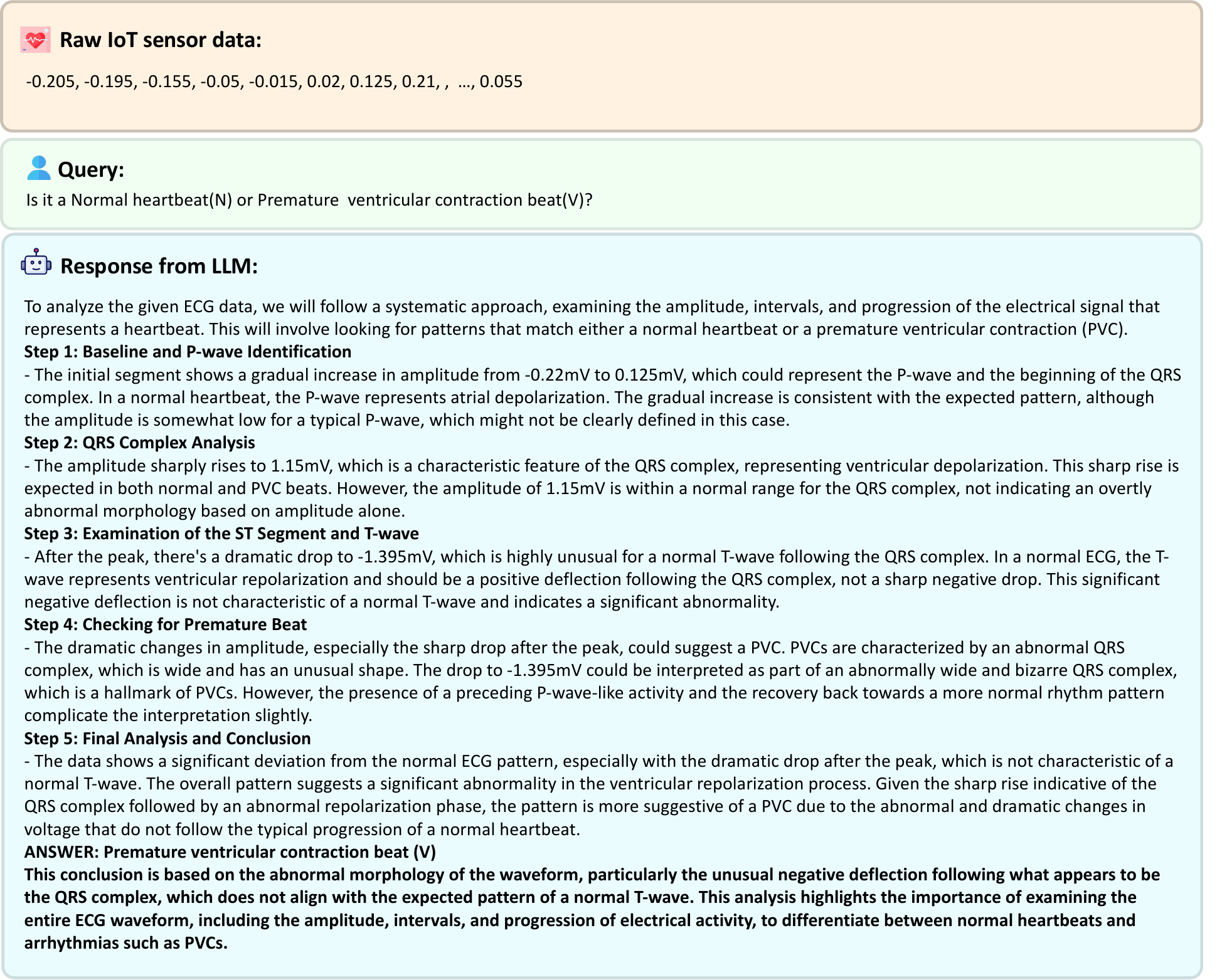}
\end{center}
\caption{Response example of LLM for heartbeat anomaly detection.}
\label{fig: heartbeat detection}
\end{figure*}

\clearpage
\begin{figure*}[b]
\begin{center}
\includegraphics[width=1\textwidth]{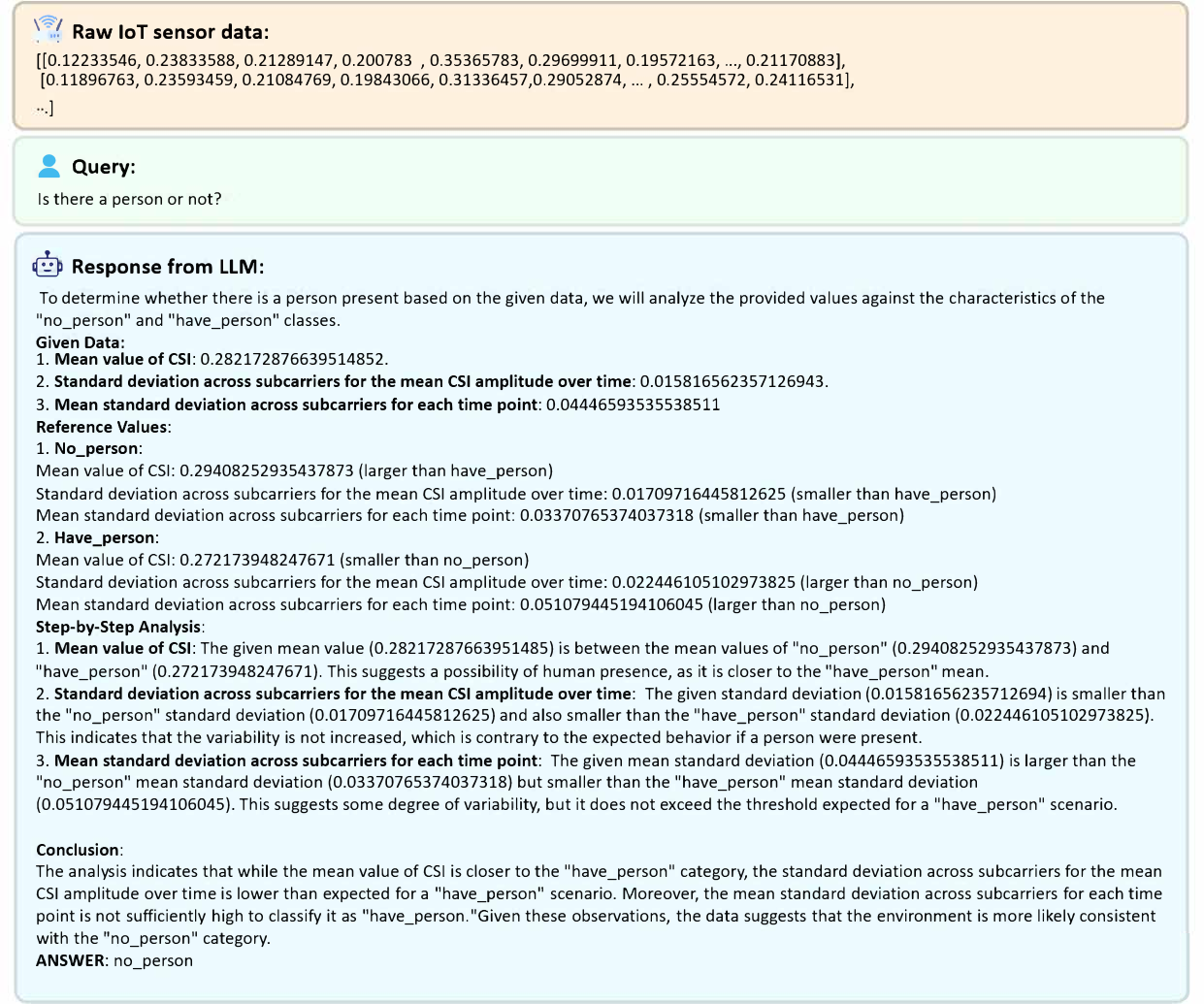}
\end{center}
\caption{Response example of LLM for WiFi-based human sensing.}
\label{fig: human sensing}
\end{figure*}

\clearpage
\begin{figure*}[b]
\begin{center}
\includegraphics[width=1\textwidth]{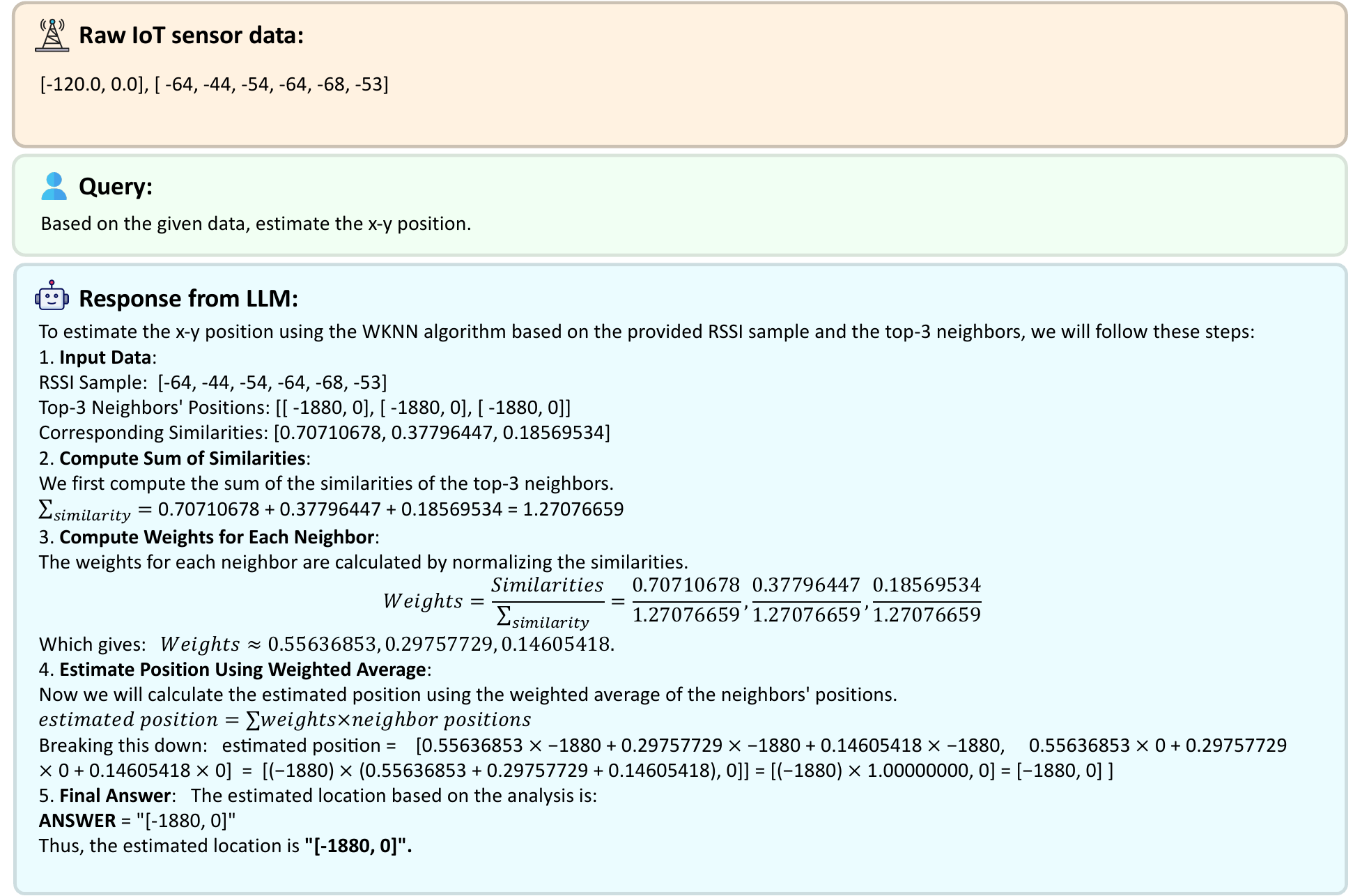}
\end{center}
\caption{Response example of LLM for WiFi-based indoor localization.}
\label{fig: localization}
\end{figure*}

\bigskip


\newpage

\end{document}